\pdfoutput=1

\documentclass[11pt]{article}

\usepackage{EMNLP2022}

\usepackage{times}
\usepackage{latexsym}

\usepackage[T1]{fontenc}

\usepackage[utf8]{inputenc}

\usepackage{microtype}

\usepackage{inconsolata}

\usepackage[pdftex]{graphicx}
\usepackage{multirow}
\usepackage{amsfonts}
\usepackage{amssymb}
\usepackage{amsmath}
\usepackage{booktabs}
\usepackage{xcolor}
\usepackage{bm}
\usepackage{numprint}
\usepackage{forest}
\usepackage{tikz-qtree}
\usepackage{numprint}
\usepackage{enumitem}
\newcommand{\revision}[1]{\textcolor[rgb]{0,0,0}{#1}}

\DeclareMathOperator*{\argmax}{arg\,max}

\title{Training a Grammar Inducer by Watching  Millions of Instructional YouTube Videos}
\title{Learning a Grammar Inducer from Massive Uncurated \\ Instructional Videos}

\author{
Songyang Zhang$^1$\Thanks{~This work was done when Songyang Zhang was an intern at Tencent AI Lab.}, Linfeng Song$^2$, Lifeng Jin$^2$, Haitao Mi$^2$, Kun Xu$^2$, \\
\textbf{Dong Yu}$^2$ and \textbf{Jiebo Luo}$^1$\\
$^1$University of Rochester, Rochester, NY, USA\\ 
\texttt{szhang83@ur.rochester.edu}, \texttt{jluo@cs.rochester.edu}\\
$^2$Tencent AI Lab, Bellevue, WA, USA \\ 
\texttt{\{lfsong,lifengjin,haitaomi,kxkunxu,dyu\}@tencent.com}\\ 
}

\begin{document}
\maketitle
\begin{abstract}
Video-aided grammar induction aims to leverage video information for finding more accurate syntactic grammars for accompanying text.
While previous work focuses on building systems for inducing grammars on text that are well-aligned with video content, we investigate the scenario, in which text and video are only in loose correspondence. Such data can be found in abundance online, and the weak correspondence is similar to the indeterminacy problem studied in language acquisition.
Furthermore, we build a new model that can better learn video-span correlation without manually designed features adopted by previous work.
Experiments show that our model trained only on large-scale YouTube data with no text-video alignment reports strong and robust performances across three unseen datasets, despite domain shift and noisy label issues.
Furthermore our model yields higher F1 scores than the previous state-of-the-art systems trained on in-domain data.
\end{abstract}

\section{Introduction}

Grammar induction is a fundamental and long-lasting \cite{lari1990estimation,clark2001unsupervised,klein2002generative}
problem in computational linguistics, which aims to find hierarchical syntactic structures from plain sentences.
Unlike supervised methods \cite{charniak2000maximum,collins2003head,petrov2007improved,zhang2011syntactic,cross2016span,kitaev2018constituency} that require human annotated treebanks, \textit{e.g.}, Penn Treebank \cite{marcus1993building}, grammar inducers do not rely on any human annotations for training.
Grammar induction is attractive since annotating syntactic trees by human language experts is expensive and time consuming, while the current treebanks are limited to several major languages and domains.

Recently, deep learning models have achieved remarkable success across
NLP tasks, and  neural models have been designed  \cite{shen2018ordered,shen2018neural,kim2019compound,kim2019unsupervised,jin2018unsupervised} for grammar induction, which greatly advanced model performance on induction with raw text.
Recent efforts have started to consider other useful information from multiple modalities, such as images \cite{shi2019visually,jin2020grounded} and videos \cite{zhang2021video}.
Specifically, \citet{zhang2021video} show that multi-modal information (e.g. motion, sound and objects) from videos can significantly improve the induction accuracy on verb and noun phrases.
Such work uses curated multi-modal data publicly available on the web, which all assume that the meaning of a sentence needs to be identical (e.g., being a caption) to the corresponding video or image.
This assumption limits usable data to several small-scale benchmarks \cite{lin2014microsoft,xu2016msr,hendricks17iccv} with expensive human annotations on image/video captions.

The noisy correspondence between form and meaning is one of the main research questions in language acquisition \cite{akhtar1999,gentner2001individuation,DOMINEY2004121}, where different proposals attempt to address this indeterminacy faced by children. There has been computational work incorporating such indeterminacy into their models \cite{yu2013grounded,NEURIPS2021_f5e62af8}. For modeling empirical grammar learning with multi-modal inputs, two important questions still remain open: 
1) \emph{how can a grammar inducer benefit from large-scale multi-media data (e.g., YouTube videos) with noisy text-to-video correspondence}?  and 
2) \emph{how can a grammar inducer show robust performances across multiple domains and datasets}?
By using data with only weak cross-modal correspondence, such as YouTube videos and their automatically generated subtitles, we allow the computational models to face a similar indeterminacy problem, and examine how indeterminacy interacts with data size to influence learning behavior and performance of the induction models.

In this paper, we conduct the first investigation on both questions.
Specifically, we collect 2.4 million video clips and the corresponding subtitles from instructional YouTube videos (HowTo100M \citealt{miech2019howto100m}) to train multi-modal grammar inducers, instead of using the training data from a benchmark where text and video are in alignment.
We then propose a novel model, named Pre-Trained Compound Probabilistic Context-Free Grammars (PTC-PCFG), that extends previous work \cite{shi2019visually,zhang2021video} by incorporating a video-span matching loss term into the Compound PCFG \cite{kim2019compound} model.
To better capture the video-span correlation, it leverages CLIP \cite{miech2020end}, a state-of-the-art model pretrained on video subtitle retrieval, as the encoders for both video and text.
Compared with previous work \cite{zhang2021video} that independently extracts features from each modality before merging them using a simple Transformer \cite{vaswani2017attention} encoder, the encoders of our model have been pretrained to merge such multi-modal information, and no human efforts are needed to select useful modalities from the full set.

Experiments on three benchmarks show that our model, which is trained on noisy YouTube video clips and no data from these benchmarks, produces substantial gains over the previous state-of-the-art system \cite{zhang2021video} trained on in-domain video clips with human annotated captions.
Furthermore, our model demonstrates robust performances across all three datasets.
We suggest the limitations of our model and future directions for improvements through analysis and discussions.
Code will be released upon paper acceptance.

In summary, the main contributions are:
\begin{itemize}[leftmargin=*]
    \item We are the first to study training a grammar inducer with massive general-domain noisy video clips instead of benchmark data, introducing the indeterminacy problem to the induction model.
    
    \item We propose PTC-PCFG, a novel model for unsupervised grammar induction. 
    It is simpler in design than previous models and can better capture the video-text matching information.
        
    
    \item Trained only on noisy YouTube videos without finetuning on benchmark data, PTC-PCFG reports stronger performances than previous models trained on benchmark data across three benchmarks.
\end{itemize}

\section{Background and Motivation}


\subsection{Compound PCFGs}
\label{sec:cpcfg}
A PCFG model in Chomsky Normal Form can be defined as a tuple of 6 terms $(\mathcal{S}, \mathcal{N}, \mathcal{P}, \Sigma, \mathcal{R}, \Pi)$, where they correspond to the start symbol, the sets of non-terminals, pre-terminals, terminals, production rules and their probabilities.
Given pre-defined numbers of non-terminals and pre-terminals, a PCFG induction model tries to estimate
the probabilities for all production rules.

The compound PCFG (C-PCFG) model \cite{kim2019compound} adopts a mixture of PCFGs.
Instead of a corpus-level prior used in previous work \cite{kurihara2006variational,johnson2007bayesian,wang2013collapsed,jin2018unsupervised}, C-PCFG imposes a sentence-specific prior on the distribution of possible PCFGs.
Specifically in the generative story, the probability $\pi_r$ for production rule $r$ is estimated by model $g$ that assigns a latent variable $\mathbf{z}$ for each sentence $\sigma$, and $\mathbf{z}$ is drawn from a prior distribution:
\begin{equation}
    \pi_r = g(r, \mathbf{z}; \theta), ~~~~~ \mathbf{z} \sim p(\mathbf{z}).
\end{equation}
where $\theta$ represents the model parameters.
The probabilities for all three types of CFG rules are defined as follows:
\begin{equation}
\begin{aligned} 
    \pi_{S\rightarrow A}&=\frac{\exp(\mathbf{u}_A^\top f_s([\mathbf{w}_S;\mathbf{z}]))}{\sum_{A'\in \mathcal{N}} \exp(\mathbf{u}_{A'}^\top f_s([\mathbf{w}_S;\mathbf{z}]))}, \\
    \pi_{A\rightarrow BC}&=\frac{\exp(\mathbf{u}_{BC}^\top [\mathbf{w}_A;\mathbf{z}])}{\sum_{B',C'\in \mathcal{N}\cup \mathcal{P}} \exp(\mathbf{u}_{B'C'}^\top [\mathbf{w}_A;\mathbf{z}]))},\\
    \pi_{T\rightarrow w}&=\frac{\exp(\mathbf{u}_w^\top f_t([\mathbf{w}_T;\mathbf{z}]))}{\sum_{w'\in \Sigma} \exp(\mathbf{u}_{w'}^\top f_t([\mathbf{w}_T;\mathbf{z}]))},
\end{aligned}
\end{equation}
where $A\in\mathcal{N}$, $B$ and $C\in\mathcal{N}\cup\mathcal{P}$, $T\in\mathcal{P}$, $w\in \Sigma$. 
Both $\mathbf{w}$ and $\mathbf{u}$ are dense vectors representing words and all types of non-terminals, and $f_s$ and $f_t$ are neural encoding functions.

Optimizing the C-PCFG model involves maximizing the marginal likelihood $p(\sigma)$ of each training sentence $\sigma$ for all possible $\mathbf{z}$:
\begin{equation}
    \log p_\theta(\sigma) = \log \int_{\mathbf{z}} \sum_{t\in \mathcal{T_G}(\sigma)} p_\theta(t|\mathbf{z}) p(\mathbf{z}) d\mathbf{z}
\end{equation}
where $\mathcal{T_G}(\sigma)$ indicates all possible parsing trees for sentence $\sigma$.
Since computing the integral over $\mathbf{z}$ is intractable, this objective is optimized by maximizing its evidence lower bound ELBO($\sigma$; $\phi$, $\theta$):
\begin{equation}
\begin{split}
        \text{ELBO}(\sigma; \phi, \theta) = \mathbb{E}_{q_\phi(\mathbf{z}|\sigma)}[\log p_\theta(\sigma|\mathbf{z})] \\
        -\text{KL}[q_\phi(\mathbf{z}|\sigma)||p(\mathbf{z})],
\end{split}
\end{equation}
where $q_\phi(\mathbf{z}|\sigma)$ is the variational posterior calculated by another neural network with parameters $\phi$.
Given a sampled $\mathbf{z}$, the log-likelihood term $\log p_\theta(\sigma|\mathbf{z})$ is calculated via the inside algorithm.
The KL term can be computed analytically when both the prior $p(\mathbf{z})$ and the variational posterior $q_\phi(\mathbf{z}|\sigma)$ are Gaussian \cite{kingma2013auto}.


\subsection{Multi-Modal Compound PCFGs}
Multi-Modal Compound PCFGs (MMC-PCFG)~\cite{zhang2021video} extends C-PCFG with a model to match a video $v$ with a span $c$ in a parse tree $t$ of a sentence $\sigma$.
It extracts $M$ visual and audio features from a video $v$ and encodes them via a multi-modal transformer~\cite{gabeur2020multi}, denoted as \revision{$\bm{\Psi}=\{\bm{\psi}^i\}_{i=1}^M$}.
The word representation $\mathbf{h}_i$ of the $i$th word is computed by BiLSTM.
Given a particular span $c=w_i,\dotsc, w_j$, its representation $\mathbf{c}$ is the weighted sum of all label-specific span representations:
\begin{align}
    \mathbf{c} 
    &=  \sum_{k=1}^{|\mathcal{N}|} p(k|c, \sigma) f_k \left(\frac{1}{j-i+1} \sum_{l=i}^{j} \mathbf{h}_l\right),
\label{eq:mm_span}
\end{align}
where $\{p(k|c, \sigma)|1\leq k \leq |\mathcal{N}|\}$ are the phrasal label probabilities of span $c$.
The representation of a span $\mathbf{c}$  is then correspondingly projected to $M$ separate embeddings via gated embedding~\cite{miech2018learning}, denoted as \revision{$\bm{\Xi}=\{\bm{\xi}^i\}_{i=1}^M$}.
Finally the video-text matching loss is defined as a sum over all video-span matching losses weighted by the marginal probability of a span from the parser:
\begin{equation}
    s_{mm}(v, \sigma) = \sum_{c\in\sigma} p(c|\sigma) h_{mm}(\bm{\Xi},\bm{\Psi}),
\end{equation}
where $h_{mm}(\bm{\Xi},\bm{\Psi})$ is a hinge loss measuring the distances from video $v$ to the matched and unmatched (\textit{i.e.} span from another sentence) span $c$ and $c^\prime$ and the distances from span $c$ to the matched and unmatched (\textit{i.e.} another video) video $v$ and $v^\prime$:
\begin{align}
    \omega_i(&\mathbf{c})=\frac{\exp(\mathbf{u}_i^\top\mathbf{c})}{\sum_{j=1}^M\exp(\mathbf{u}_j^\top\mathbf{c})},\\
    o(\bm{\Xi},&\bm{\Psi})=\sum_{i=1}^M\omega_i(\mathbf{c})\mathrm{cos}(\bm{\xi}^i,\bm{\psi}^i), \\
    h_{mm}(\bm{\Xi},&\bm{\Psi}) = \mathbb{E}_{{c}^\prime}[o(\bm{\Xi}^\prime, \bm{\Psi}) - o(\bm{\Xi}, \bm{\Psi})) + \epsilon ]_+ \nonumber \\ 
    &+ \mathbb{E}_{v^\prime}[o(\bm{\Xi}, \bm{\Psi}^\prime) - o(\bm{\Xi}, \bm{\Psi}) + \epsilon]_+,
\end{align}
where \revision{$\bm{\Xi}^\prime$ is a set of unmatched span expert embeddings of $\bm{\Psi}$, $\bm{\Psi}^\prime$ is a set of unmatched video representations of $\bm{\Xi}$,} $\epsilon$ is a positive margin, $[\cdot]_+ = max(0, \cdot)$, $\{\mathbf{u}_i\}_{i=1}^M$ are learned weights, and the expectations are approximated with one sample drawn from the training data.
During training, both ELBO and the video-text matching loss are jointly optimized.

\subsection{Limitation and Motivation}
Existing work on multi-modal grammar induction aims at leveraging strict correspondence between image/video and text for information about syntactic categories and structures of the words and spans in the text. However, such datasets are 
expensive to annotate. Besides, the ambiguous correspondence between language and real-world context, observed in language acquisition, is not really reflected in such training setups.


As a result, we believe that the previous work fails to answer the following important questions: 1) how well a grammar inducer would perform when it is trained only on noisy multi-media data; 2) how the scale of training data would affect the performance and cross-domain robustness?



\begin{figure*}[t!]
    \centering
    \includegraphics[width=0.95\textwidth]{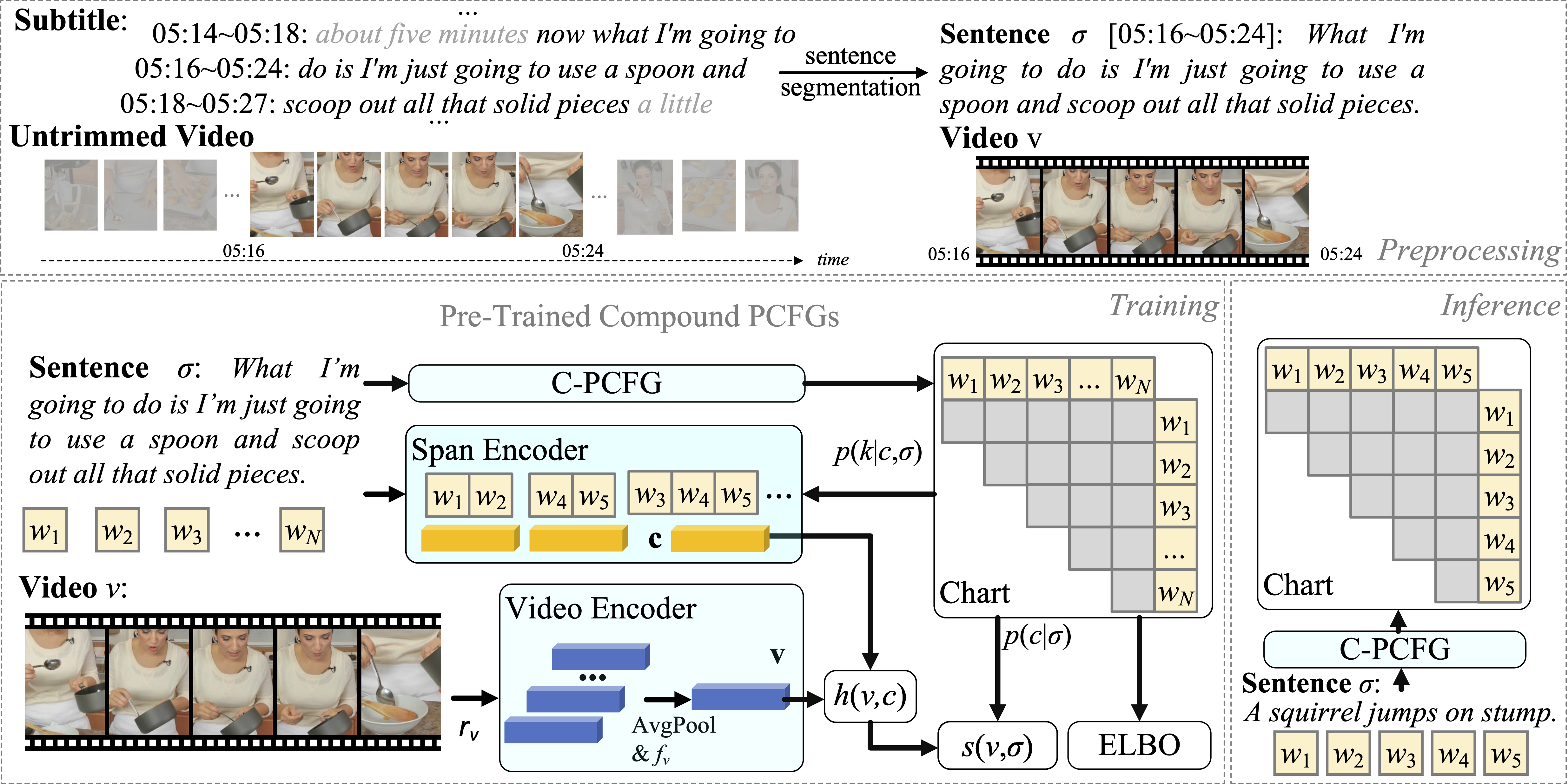}
    \caption{The pipeline of our approach.}
    \label{fig:framework}
\end{figure*}

\section{Training a Grammar Inducer with Massive YouTube Videos}

We make the first investigation into the above questions by leveraging massive video clips from instructional YouTube videos to train our grammar inducer.
Different from the benchmark data used by previous work, the YouTube video clips do not contain paired sentences.
This section will first introduce the method for generating noisy training instances (video clip and sentence pairs) from YouTube videos (\S \ref{sec:data_process}), before describing a novel grammar induction model (\S \ref{sec:model}) with pre-trained text and video encoders.

\subsection{Harvesting Training Instances from YouTube Videos}
\label{sec:data_process}

Given a YouTube video, we would like to generate a set of video clip and subtitle pairs $\Omega=\{(v, \sigma)\}$, where each subtitle $\sigma$ is a complete sentence and is aligned in time with its paired video clip $v$.
To this end, the YouTube API is chosen to obtain all subtitles of the video.
But, our observation finds that most obtained subtitles are not complete sentences, and in some cases, a complete sentence can last for several continuous video fragments. 
Meanwhile, they do not contain any punctuation, which is a key factor for sentence segmentation.
As shown in the top part of Figure \ref{fig:framework}, we design an algorithm that takes the following steps to find each complete sentence and its corresponding video clip.

\textbf{Sentence segmentation.}
In the first step, we try to find complete sentences from the subtitles.
We first concatenate all subtitles from the same video are concatenated into a very long sequence of tokens.
Next, a punctuation restoration model\footnote{\revision{We manually punctuate subtitles from $10$ videos randomly selected from HowTo100M, which contains $461$ sentences after annotation. The punctuation restoration model has an overall F1 score of $74.1\%$ with the manual labels.}}~\cite{tilk2016} is adopted to insert punctuation into the sequence.
Lastly, sentences are segmented based on certain punctuation (\textit{e.g.}, ``\textit{.}'', ``\textit{?}'', ``\textit{!}'').

\textbf{Video clip extraction.}
In the second step, we trim the corresponding video clips.
Each raw subtitle contains its start and an end times. 
We assume each word within the raw subtitle occupies equal time and record the start and end times for each word.
After that, given a complete sentence $\sigma=w_1,w_2,...,w_N$, we use the start time of its first word $w_1$ and the end time of its last word $w_N$ as the start and end times of $\sigma$. Lastly, we segment a complete sentence $\sigma$'s corresponding video clip $v$ based on its start and end times.



\subsection{Model: Pre-Trained Compound PCFGs}
\label{sec:model}

After harvesting large-scale sentence and video pairs, the next step is to build a strong grammar induction model that can benefit from them.
In this section, we introduce our Pre-Trained Compound PCFGs (PTC-PCFG) model for unsupervised grammar induction.
As shown in the lower part of Figure~\ref{fig:framework}, the PTC-PCFG model composes of a video encoder, a span encoder and a parsing model.
Both the video encoder and the span encoder are initialized from the MIL-NCE model~\cite{miech2020end}, a pre-trained video-text matching model that takes a simple design and has shown superior zero-shot results on many video understanding tasks, such as video retrieval, video question answering, \textit{etc}.
We first introduce the pre-trained video and span encoders, before covering the training and inference details of PTC-PCFG.

\noindent\textbf{Video encoding.}
The first step is to encode a video $v$ to its representation $\mathbf{v}$.
To do this, we first segment $v$ into small video clips, where each video clip $v_i$ consists of $T$ frames. Following~\citet{zhang2021video}, we sample $L$ video clips with equal interval for efficiency.
We use the video encoder from the MIL-NCE model~\cite{miech2020end} as our video encoder and only fine-tune its last fully connected layer $f_{v}$ for efficiency.
In more detail, for each sampled video clip, we pre-compute the input of $f^{v}$ as its representation, denoted as $\{\mathbf{h}^v_i\}_{i=1}^L$. Then we feed them into $f^v$ and average the output as its representation $\mathbf{v}$, denoted as,
\begin{align}
    \mathbf{v} &= \texttt{AvgPool}(\{f^{v}(\mathbf{h}^v_i)\}_{i=1}^L),
\end{align}
where $\texttt{AvgPool}$ indicates average pooling.



\noindent\textbf{Span encoding.}
The next step is to compute a span representation $\mathbf{c}$ for each particular span $c=w_i,\dots,w_j$ $(1\leq i< j \leq N)$ in sentence $\sigma=w_1,w_2,\dots,w_N$.
The pre-trained text encoder of MIL-NCE consists of a word embedding layer and two stacked fully connected layers, $f^c_0$ and $f^c_1$.
Motivated by~\citet{zhao2020visually,zhang2021video}, we expect to learn $|\mathcal{N}|$ different span representations, each is specified for one non-terminal node.
However, directly applying the pre-trained text encoder is not feasible, since it has only one output layer $f^c_1$.
Therefore, we duplicate $f^c_1$ for $|\mathcal{N}|$ times, denoted as $\{f^c_{k}\}_{k=1}^{|\mathcal{N}|}$, and compose $|\mathcal{N}|$ label-specific output layers.
In more detail, we first encode each word $w_i$ with the word embedding layer, denoted as $\mathbf{h}^c_i$. 
Then we feed the word embeddings to $f^c_0$, ReLU, maximum pooling and each label-specific output layer sequentially.
we also compute the probabilities of its phrasal labels $\{p(k|c, \sigma) |1\le k \le |\mathcal{N}|\}$, as illustrated in Section~\ref{sec:cpcfg}.
Lastly, the span representation $\mathbf{c}$ is the sum of all label-specific span representations weighted by the probabilities we predicted, denoted as:
\begin{equation}
\begin{split}
    \mathbf{\tau} &= \texttt{MaxPool}(\texttt{ReLU}(f^c_0(\mathbf{h}^c_i))) \\
    \mathbf{c} &=  \sum_{k=1}^{|\mathcal{N}|} p(k|c, \sigma) f^c_k(\mathbf{\tau}),
\label{eq:pt_span}
\end{split}
\end{equation}
where \texttt{MaxPool} is a maximum pooling operation and \texttt{ReLU} is a ReLU activation function.

\noindent\textbf{Training.}
As shown in lower left of Figure~\ref{fig:framework}, we optimize both the video-text matching loss and evidence lower bound during training.
We first compute the similarity between a video clip $v$ and a particular span $c$ via dot product and then compute a triplet hinge loss as following,
\begin{align}
    h(v,c)&=\mathbb{E}_{c^\prime}[\mathbf{c^\prime}\cdot\mathbf{v}-\mathbf{c}\cdot\mathbf{v}+\epsilon]_{+} \nonumber\\
    &+\mathbb{E}_{v^\prime}[\mathbf{c}\cdot\mathbf{v^\prime}-\mathbf{c}\cdot\mathbf{v}+\epsilon]_{+},
\end{align}
where $\epsilon$ is a positive margin, $[\cdot]_+ = max(0, \cdot)$, $v^\prime$ is a clip from a different video and $c^\prime$ is a span from a different sentence.
The video-text matching loss is correspondingly defined as,
\begin{align}
    s(v,\sigma) = \Sigma_{c\in \sigma} p(c|\sigma)h(v,c),
\end{align}
where $p(c|\sigma)$ is the probability of a particular span $c$ being a syntactic phrase.
Finally, the overall loss function is composed by the $\mathrm{ELBO}$ and the video-text matching loss:
\begin{equation}
    \mathcal{L}(\phi,\theta)=\sum_{(v,\sigma)\in \Omega}-\mathrm{ELBO}(\sigma;\phi,\theta)+\alpha s_{}(v,\sigma),
\end{equation}
where $\alpha$ is a constant balancing these two terms.

\noindent\textbf{Inference.}
During inference, given a sentence $\sigma$, we predict the most likely tree $t^*$ without accessing videos, as shown in the lower right of Figure~\ref{fig:framework}. Since computing the integral over $\mathbf{z}$ is intractable, we estimate $t^*$ with the following approximation,
\begin{equation}
\begin{aligned}
    t^*&=\argmax_t\int_{\mathbf{z}} p_\theta (t|\mathbf{z}) p_\theta(\mathbf{z}|\sigma) d \mathbf{z}\\
    &\approx \argmax_t p_\theta (t|\sigma,\bm{\mu}_{\phi}(\sigma)),
\end{aligned}
\end{equation}
where $\bm{\mu}_{\phi}(\sigma)$ is the mean vector of the variational posterior $q_\phi(\mathbf{z}|\sigma)$, and $t^*$ is obtained by the CYK algo.~\cite{cocke1969programming,younger1967recognition,kasami1966efficient}.

\section{Experiments}


\subsection{Datasets}

Following previous work, we evaluate all systems on three benchmarks (i.e., DiDeMo, YouCook2 and MSRVTT). Instead of training on benchmark data, our models are trained on the data harvested from HowTo100M dataset. Below shows more details about these datasets:


\noindent\textbf{DiDeMo}~\cite{hendricks17iccv} contains $10$k unedited personal Flickr videos. Each video is associated with roughly $3$-$5$ video-sentence pairs. There are $\numprint{32994}$, $\numprint{4180}$ and $4021$ video pairs in the training, validation and testing sets.

\noindent\textbf{YouCook2}~\cite{ZhXuCoAAAI18} contains $2000$ long untrimmed YouTube videos from $89$ cooking recipes. The procedure steps for each video are annotated with temporal boundaries and described by imperative English sentences. 
There are $\numprint{8913}$, $\numprint{969}$ and $\numprint{3310}$ video-sentence pairs in the training, validation and testing sets.

\noindent\textbf{MSRVTT}~\cite{xu2016msr} contains $10$k generic YouTube videos accompanied by $200$k captions annotated by paid human workers. 
There are $\numprint{130260}$, $\numprint{9940}$ and $\numprint{59794}$ video-sentence pairs in the training, validation and testing sets.

\noindent\textbf{HowTo100M}~\cite{miech2019howto100m} is a large-scale dataset of $136$ million video clips sourced from $1.22$M narrated instructional web videos depicting humans performing more than $23$k different visual tasks.
Noted that there are $404$ videos in HowTo100M exists in YouCook2, we exclude these videos during training.

\subsection{Evaluation}

We discard punctuation, lowercase all words, replace numbers with a special token and ignore trivial single-word and sentence-level spans during testing following~\citet{kim2019compound}. 
Besides, we follow previous work \cite{shi2019visually,zhang2021video} by using a state-of-the-art constituency parser (Benepar~\citealt{kitaev-etal-2019-multilingual}) to obtain the reference trees for evaluation\footnote{\revision{For each dataset, we randomly select $50$ sentences and manually label their constituency parse trees. Benepar has S-F1 scores of $98.1\%$ (DiDeMo), $97.2\%$ (YouCook2) and $98.1\%$ (MSRVTT) with manual labels.}}. \revision{Following~\citet{shi2020role,zhang2021video}, all models are run $5$ times for $1$ epoch with different random seeds. For each model, we report the averaged sentence-level F1 (S-F1) and corpus-level F1 (C-F1) of its runs on each testing set.}

\subsection{Implementation Details}
\label{sec:implementation_details}

We use Spacy~\footnote{\url{https://spacy.io/}} for tokenization and keep sentences with fewer than $40$ words for training due to the limited computational resources.
Each video is decoded at $16$ fps and $L=8$ video clips are sampled in total, where each clip contains $T=16$ frames.
We train baseline models, C-PCFG and MMC-PCFG with the same hyper-parameters suggested by ~\citet{kim2019compound} and \citet{zhang2021video}.
The parsing model of PTC-PCFG has the same hyper-parameter setting as C-PCFG and MMC-PCFG (Please refer their papers for details).
The constant $\alpha$ is set to $1$.
We select the top $\numprint{20000}$ most common words in HowTo100M as vocabulary for all datasets.
All baseline methods and ours are optimized by Adam~\cite{kingma2014adam} with a learning rate of $0.001$, $\beta_1=0.75$ and $\beta_2=0.999$. 
All parameters (except the video-text matching model in PTC-PCFG) are initialized with Xavier uniform initializer~\cite{glorot2010understanding}. 
All our models in experiments are trained for $1$ epoch with batch size of $32$, without finetuning on the target dataset.

\subsection{Main Results}

\begin{figure}[t!]
    \centering
    \includegraphics[width=0.45\textwidth]{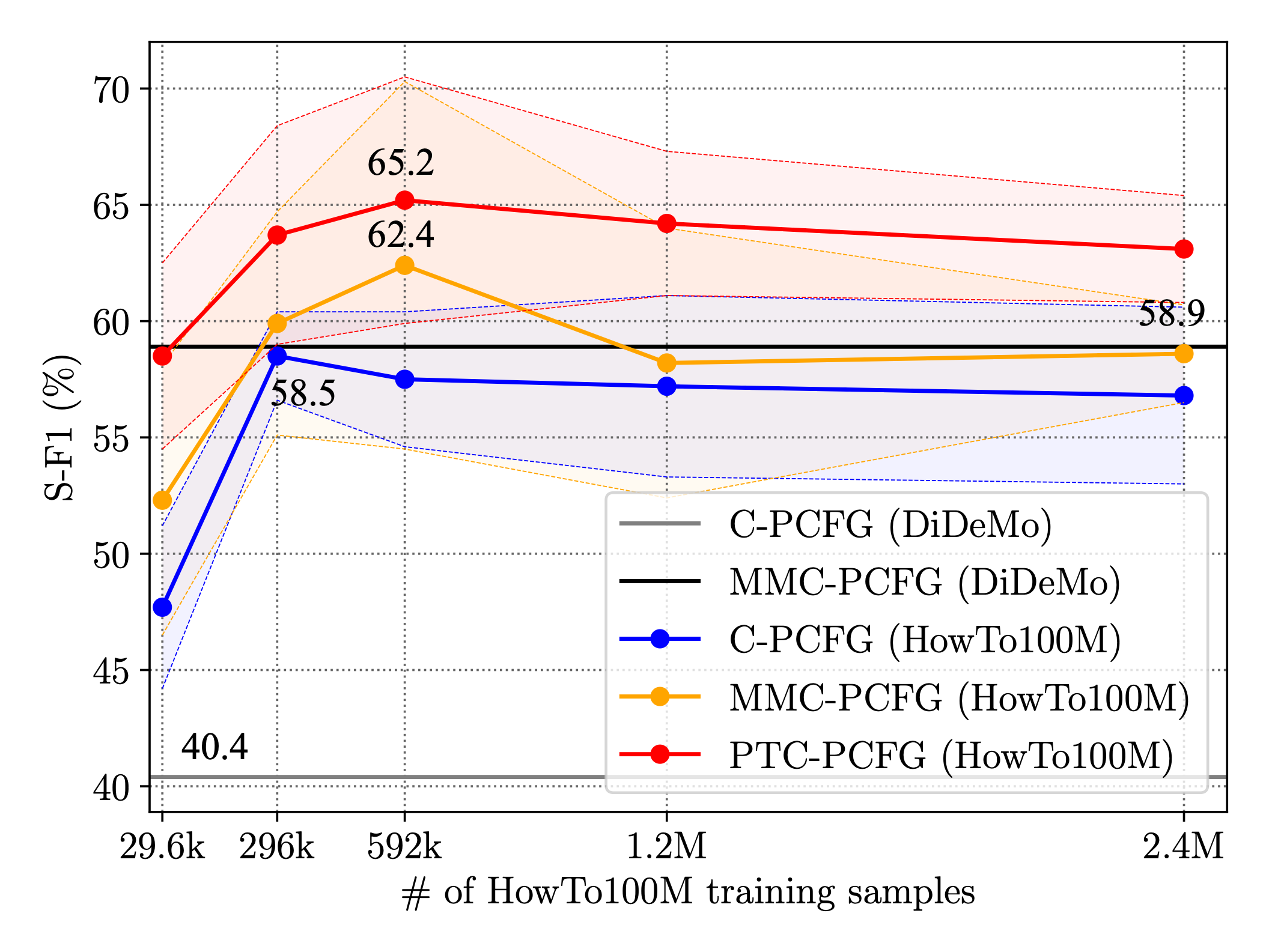}
    \caption{Performance Comparison on DiDeMo. The doted lines and their enclosed area represent the mean and variance of each model trained on HowTo100M at different scales. We mark the highest average S-F1 achieved by each method with numbers. The remaining figures follow the same notations.}
    \label{fig:didemo_scale}
\end{figure}

\begin{figure}[t!]
    \centering
    \includegraphics[width=0.45\textwidth]{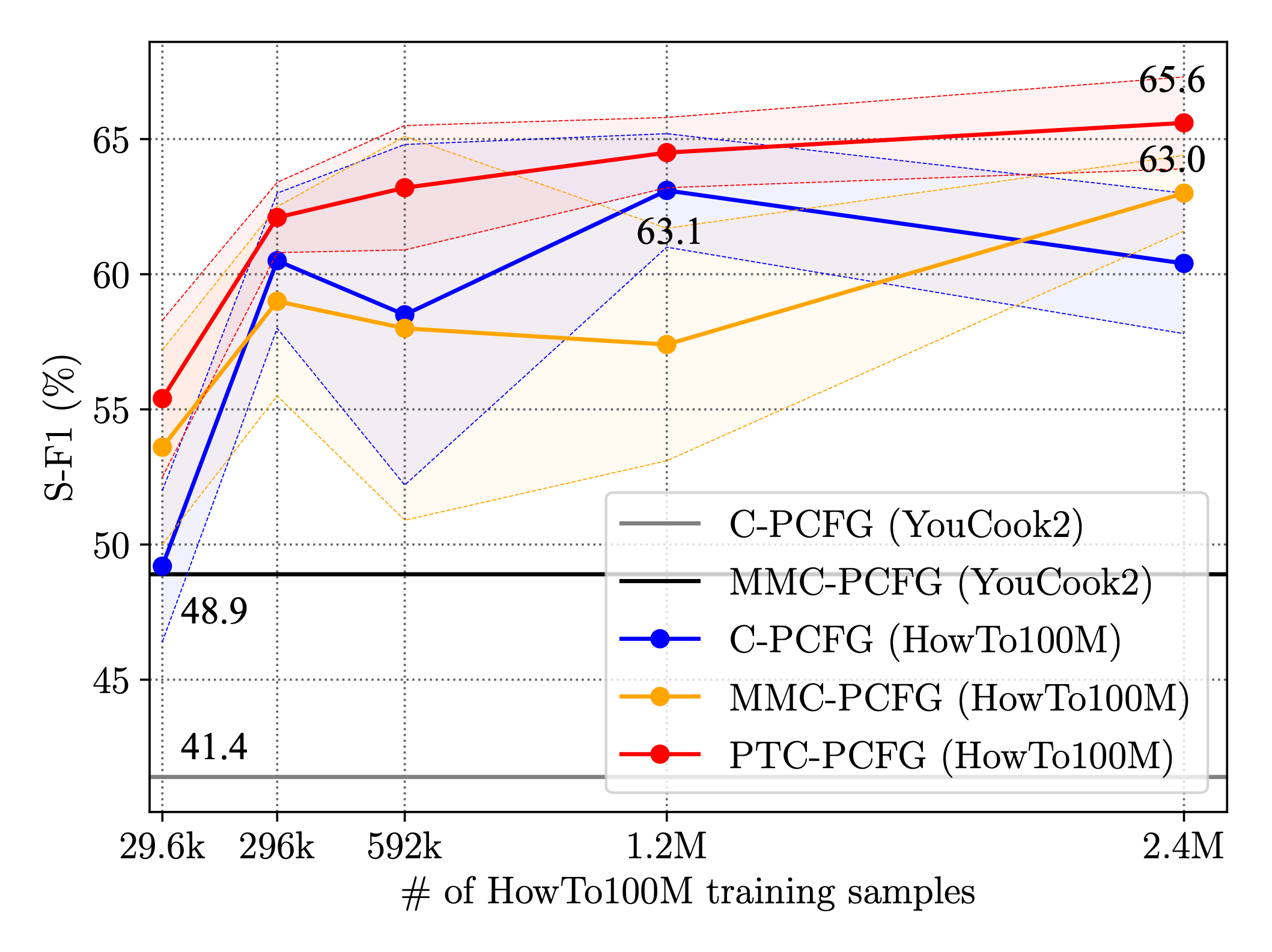}
    \caption{Performance Comparison on YouCook2.}
    \label{fig:youcook2_scale}
\end{figure}

\begin{figure}[t!]
    \centering
    \includegraphics[width=0.45\textwidth]{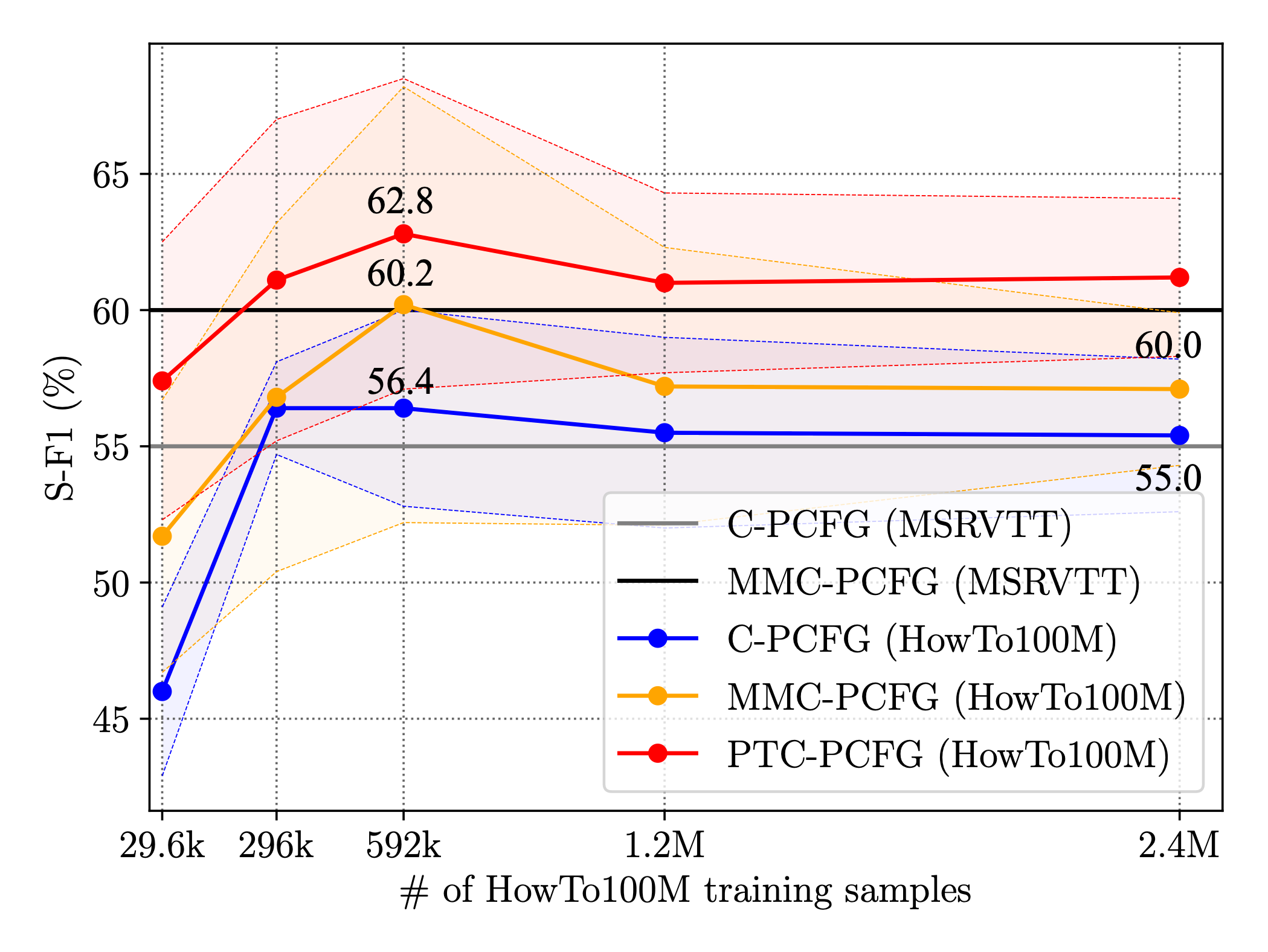}
    \caption{Performance Comparison on MSRVTT.}
    \label{fig:msrvtt_scale}
\end{figure}

Figure~\ref{fig:didemo_scale}-\ref{fig:msrvtt_scale} compare our proposed PTC-PCFG approach with recently proposed state-of-the-art models: C-PCFG~\cite{kim2019compound} and MMC-PCFG\footnote{Since audios are removed by HowTo100M authors, we implement MMC-PCFG with video features only, including object features(ResNeXt, SENet), action features (I3D, R2P1D, S3DG), scene features, OCR features and face features.}~\cite{zhang2021video}.
To pinpoint more fine-grained contributions, we also train these models on HowTo100M data.

\noindent\textbf{The effectiveness of HowTo100M.}
\revision{
We find that C-PCFG achieve better performance when they are trained with more instances from HowTo100M than the original in-domain training sets, where the largest improvements are $+18.1\%$, $+21.7\%$ and $+1.4\%$ S-F1 scores on DiDeMo, YouCook2 and MSRVTT, respectively.
}
These results indicate that grammar inducers are generally robust against the instances with noisy text-video correspondence.
As the results, learning from noisy YouTube videos can benefit model's overall performance and its generalization ability across multiple domains.

\noindent\textbf{The effectiveness of PTC-PCFG.}
Comparing 
C-PCFG, MMC-PCFG and PTC-PCFG trained on different amount of HowTo100M data, we found that PTC-PCFG achieves the best performances in all three datasets.
It can further improve S-F1 to \revision{$+6.3\%$} on DiDeMo, \revision{$+16.7\%$} on YouCook2 and \revision{$+2.8\%$} on MSRVTT. 
This demonstrates the effectiveness of the PTC-PCFG model.
In particular, utilizing the video and span encoders pre-trained on a relevant tasks (\textit{e.g.}, video retrieval) can benefit unsupervised grammar induction.

\noindent\textbf{Performance comparison over data scale.}
On DiDeMo and MSRVTT, we observe that PTC-PCFG achieves the best performance with \revision{$592$}k HowTo100M training samples, and further increasing the number of training instances does not improve the parsing performance on these two datasets.
In contrast, the performance gain of PTC-PCFG on YouCook2 further increases with increasing training data.
The reason can be that the domain of HowTo100M is closer to YouCook2 (both are instructional videos) than the other two datasets.
Future work includes adding data from other sources to the whole training set more domain generic.



\begin{table*}[t!]
\small
    \centering
    \caption{Performance comparison across different training set. We use HT to represent HowTo100M dataset for short, where the number in the brackets indicates the number of samples used for training.  The values highlighted by \textbf{bold} and \textit{italic} fonts indicate the top-2 methods, respectively. All numbers are shown in percentage($\%$). The remaining tables follow the same notations.}
    \begin{tabular}{cccccccc}
    \toprule
    \multirow{2}{*}{Method} & \multirow{2}{*}{Trainset} & \multicolumn{2}{c}{DiDeMo} & \multicolumn{2}{c}{YouCook2} & \multicolumn{2}{c}{MSRVTT}\\
	\cmidrule(lr){3-4}
	\cmidrule(lr){5-6}
	\cmidrule(lr){7-8}
    & & C-F1 & S-F1 & C-F1 & S-F1 & C-F1 & S-F1 \\
	\midrule
    MMC-PCFG & DiDeMo & ${55.0}_{\pm3.7 }$ & ${58.9}_{\pm3.4 }$ & $49.1_{\pm4.4}$ & $53.0_{\pm4.9}$ & $49.6_{\pm1.4}$ & $53.8_{\pm0.9}$ \\
    MMC-PCFG & YouCook2 & $40.1_{\pm4.4}$ & $44.2_{\pm4.4}$ & ${44.7}_{\pm5.2}$ & ${48.9}_{\pm5.7}$ & $34.0_{\pm6.4}$ & $37.5_{\pm6.8}$ \\ 
    MMC-PCFG & MSRVTT & $\mathit{59.4}_{\pm2.9}$ & $\mathit{62.7}_{\pm3.3}$ & $49.6_{\pm3.9}$ & $54.2_{\pm4.1}$ & $\mathit{56.0}_{\pm1.4}$ & $60.0_{\pm1.2}$ \\
    MMC-PCFG & HT(592k) & $58.5_{\pm7.3}$ & $62.4_{\pm7.9}$ & $\mathit{53.9}_{\pm6.6}$ & $\mathit{58.0}_{\pm7.1}$ & $55.1_{\pm7.0}$ & $\mathit{60.2}_{\pm8.0}$ \\
    \textbf{PTC-PCFG} & HT(592k) & $\mathbf{61.3}_{\pm3.9}$ & $\mathbf{65.2}_{\pm5.3}$ & $\mathbf{58.9}_{\pm2.5}$ & $\mathbf{63.2}_{\pm2.3}$ & $\mathbf{57.4}_{\pm4.6}$ & $\mathbf{62.8}_{\pm5.7}$ \\
    \bottomrule
    \end{tabular}
    \label{tab:cross-dataset}
\end{table*}

\subsection{Cross-dataset Evaluation}

We evaluate the robustness of models across different datasets, as shown in Table~\ref{tab:cross-dataset}.
Comparing MMC-PCFG trained on in-domain datasets (Row $1$-$3$), we can observe that MMC-PCFG trained on MSRVTT achieves the best overall performance, while MMC-PCFG trained on YouCook2 is the worst.
We believe this is due to the different number of training instances\footnote{The number of training instances in YouCook2, DiDeMo and MSRVTT are 8.9K, 32.9K and 130.2K, respectively.} and the domain gap between different datasets.
Comparing Rows $1$-$4$, we can observe that the MMC-PCFG model trained on HT(592k) (Row $4$) is the best or the second place regarding C-F1 and S-F1 compared with its variants trained on in-domain datasets (Rows $1$-$3$). 
This demonstrates that the our processed video-text training instances are abundant, rich in content and can serve for general purpose.
Comparing Rows $4$ and $5$, PTC-PCFG outperforms MMC-PCFG in both C-F1 and S-F1 in all three datasets and has smaller variance.
This demonstrate that our model can leverage pre-trained video-text matching knowledge and learn consistent grammar induction.

\begin{table*}[t!]
\small
    \centering
    \caption{Performance comparison across different video and span encoders.}
    \begin{tabular}{ccccccccc}
    \toprule
    \multicolumn{2}{c}{Video-Text Model} & \multirow{2}{*}{Trainset} & \multicolumn{2}{c}{DiDeMo} & \multicolumn{2}{c}{YouCook2} & \multicolumn{2}{c}{MSRVTT}\\
	\cmidrule(lr){1-2}
	\cmidrule(lr){4-5}
	\cmidrule(lr){6-7}
	\cmidrule(lr){8-9}
    Video Encoder& Span Encoder & & C-F1 & S-F1 & C-F1 & S-F1 & C-F1 & S-F1 \\
	\midrule
    MIL-NCE & LSTM & HT(296k) & $52.4_{\pm5.5}$ & $54.4_{\pm5.4}$ & $51.5_{\pm5.4}$ & $56.5_{\pm5.2}$ & $49.7_{\pm5.5}$ & $53.4_{\pm5.8}$ \\
    MM & LSTM & HT(296k) & $53.6_{\pm3.2}$ & $55.8_{\pm3.1}$ & $53.1_{\pm5.7}$ & $57.9_{\pm5.6}$ & $48.9_{\pm3.5}$ & $52.5_{\pm3.6}$ \\
    MIL-NCE & TinyBERT & HT(296k) & $\mathit{54.8}_{\pm5.4}$ & $\mathit{56.4}_{\pm6.0}$ & $\mathit{55.7}_{\pm4.0}$ & $\mathit{60.2}_{\pm3.5}$ & $\mathit{52.3}_{\pm4.3}$ & $\mathit{56.0}_{\pm5.0}$ \\
    MIL-NCE & MIL-NCE & HT(296k) & $\mathbf{59.5}_{\pm4.3}$ & $\mathbf{63.7}_{\pm4.7}$ & $\mathbf{57.1}_{\pm1.7}$ & $\mathbf{62.1}_{\pm1.3}$ & $\mathbf{55.7}_{\pm5.0}$ & $\mathbf{61.1}_{\pm5.9}$ \\
    CLIP & CLIP & HT(296k) & $52.9_{\pm2.3}$ & $54.9_{\pm2.6}$ & $53.3_{\pm2.2}$ & $58.9_{\pm2.1}$ & $49.1_{\pm2.6}$ & $53.0_{\pm2.9}$ \\
    \bottomrule
    \end{tabular}
    \label{tab:pretraining}
\end{table*}

\subsection{Effectiveness of Pre-Training}

In this section, we explore how different pre-trained video and text encoders can affect the parsing performance, and the results are shown in Table~\ref{tab:pretraining}.
In particular, we study different video encoders\footnote{~\revision{We list the video processing details in Appendix~\ref{appendix:preprocessing_details}.}}, including the S3D-based encoder from MIL-NCE~\cite{miech2020end} (\textit{MIL-NCE}), the multi-modal video encoder from MMC-PCFG~\cite{zhang2021video} (\textit{MM}) and the CLIP model for image-text pre-training~\cite{radford2021learning} (\textit{CLIP}).
We also investigate various text encoders, including an LSTM encoder with random initialization~\cite{zhang2021video,zhao2020visually}, a pre-trained TinyBERT~\cite{jiao2020tinybert} model, the text encoder from MIL-NCE~\cite{miech2020end}, and the text encoder from CLIP~\cite{radford2021learning}.

Comparing Rows $1$ with $2$, we can observe that MM is better than the video encoder of MIL-NCE regarding C-F1 and S-F1 on all three datasets, as MM provides more comprehensive video features.
By comparing row $1$ with $3$, we can also observe that TinyBERT, which is distilled from BERT \cite{devlin2019bert}, outperforms the randomly initialized LSTM encoder.
However, both MM and TinyBERT are independently trained only on vision or language tasks, where the vision-language correspondences are not considered during pre-training. 
Therefore, we further investigate the encoders jointly pre-trained on large scale multi-media datasets, including the video-text matching model MIL-NCE (Row $4$) and the image-text matching model CLIP (Row $5$).
We can observe that by leveraging both video and text encoders in MIL-NCE can improve the parsing performance by a large margin on all three datasets.
On the other hand, CLIP does not perform well, since it is designed for static images and other multi-modal information (e.g., motion) is ignored.

\subsection{Qualitative Analysis}

\begin{figure}
    \centering
    \includegraphics[width=0.48\textwidth]{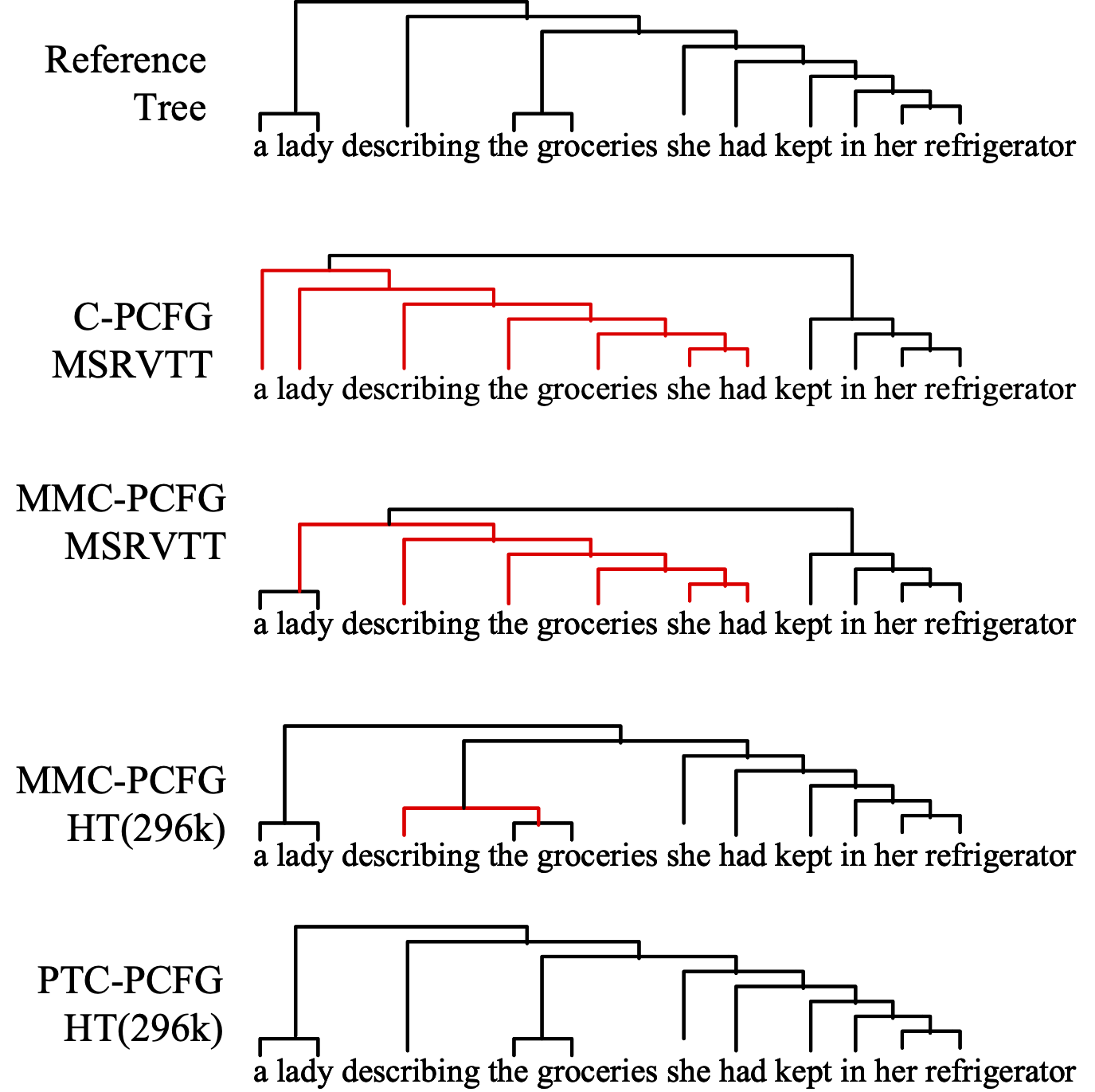}
    \caption{Parse trees predicted by different models for sentence \textit{a lady describing the groceries she had kept in her refrigerator}. The \textcolor{red}{red} line shows the difference between the predicted trees and the reference tree.}
    \label{fig:examples}
\end{figure}

In figure~\ref{fig:examples}, we visualize a parser tree predicted by the best run of C-PCFG trained on MSRVTT, MMC-PCFG trained on MSRVTT, MMC-PCFG trained on HT(296k) and PTC-PCFG trained on HT(296k), as well as its reference tree. We can observe that C-PCFG trained on MSRVTT fails at noun phrase ``\textit{a lady}'', while MMC-PCFG trained on MSRVTT succeeds. MMC-PCFG can be further improved by training on HT(296k), however, fails at noun phrase ``\textit{the groceries she had kept in her refrigerator}''. Our PTC-PCFG can leverage the pre-trained matching knowledge and make the correct prediction.

\section{Related Work}

\noindent\textbf{Grammar Induction}
 has a long and rich history in the computational linguistics.
Earlier work~\cite{shen2018neural,shen2018ordered,drozdov2019unsupervised,kim2019compound,Jin2019-us,yang-etal-2021-neural,yang-etal-2021-pcfgs} on grammar induction with pure unsupervised learning showed promising results. Instead of learning purely from text, recent work improved the parsing performance with paired images~\cite{shi2019visually,zhao2020visually} or videos~\cite{zhang2021video}. However, they are all limited to small benchmarks and specified for a few domains. In contrast, our work leverages massive noisy video-subtitle pairs from YouTube without any manual annotations.

\noindent\textbf{Video Retrieval}
has been a hot topic in the computer vision field for many years. Earlier approaches focused on model design~\cite{gabeur2020multi,zhang2019exploiting}, while more recent approaches~\cite{radford2021learning,miech2020end} focused on the pre-training on a large scale dataset and demonstrated superior zero-shot results on 
many downstream tasks.
These models are simple in design and provide representative features with less human effort in annotations. In this work, we demonstrate that unsupervised grammar induction can also benefit from the pre-trained video-text model.

\section{Conclusion}

In this paper, we have investigated how massive instructional YouTube video and subtitle pairs can improve grammar induction. We have also proposed a new model that leverages the latest advances in multi-modal pre-training to learn better video-span correlation. Experiments on three benchmarks demonstrate superior and robust performances of our model over previous systems.
We leave exploring other pre-trained video-text matching models and more publicly available data (e.g., YouTube videos from other domains and TV shows) in future work.

\section{Limitations}
Although our model faces a similar indeterminacy problem like children do, and results show that induction works even with noisy correspondence, there are a few factors which prevent this result from being directly applied to language acquisition. Our models only use instructional video and do not have the capability to interact with the world, both of which are unrealistic for human language learners. The complexity of the PCFG induction algorithm we use is cubic to the number of syntactic categories, therefore potentially limits the usefulness of larger amounts of data, where finer subcategories may be learned. Algorithms such as in \citet{yang-etal-2021-pcfgs} could be used in conjunction with multimodal inputs to examine this issue.

Following previous work, our experiments are only conducted on English video-text datasets. However, our framework is general for grammar induction in many languages. 
Since our training instances are originally collected from Internet and are uploaded by users, the dataset itself might have misinformation. 
Meanwhile, training a model on a large-scale dataset could have high cost in energy and carbon emission. We list our computational cost of our experiments in Appendix~\ref{appendix:computational_cost}.

\bibliography{reference}

\begin{thebibliography}{61}
\expandafter\ifx\csname natexlab\endcsname\relax\def\natexlab#1{#1}\fi

\bibitem[{Akhtar and Montague(1999)}]{akhtar1999}
Nameera Akhtar and Lisa Montague. 1999.
\newblock Early lexical acquisition: the role of cross-situational learning.
\newblock \emph{First Language}, 19(57):347--358.

\bibitem[{Carreira and Zisserman(2017)}]{carreira2017quo}
Joao Carreira and Andrew Zisserman. 2017.
\newblock Quo vadis, action recognition? a new model and the kinetics dataset.
\newblock In \emph{CVPR}.

\bibitem[{Charniak(2000)}]{charniak2000maximum}
Eugene Charniak. 2000.
\newblock A maximum-entropy-inspired parser.
\newblock In \emph{NAACL}.

\bibitem[{Clark(2001)}]{clark2001unsupervised}
Alexander Clark. 2001.
\newblock Unsupervised induction of stochastic context-free grammars using
  distributional clustering.
\newblock In \emph{ConLL}.

\bibitem[{Cocke(1969)}]{cocke1969programming}
John Cocke. 1969.
\newblock \emph{Programming languages and their compilers: Preliminary notes}.
\newblock New York University.

\bibitem[{Collins(2003)}]{collins2003head}
Michael Collins. 2003.
\newblock Head-driven statistical models for natural language parsing.
\newblock \emph{Computational linguistics}, 29(4):589--637.

\bibitem[{Cross and Huang(2016)}]{cross2016span}
James Cross and Liang Huang. 2016.
\newblock Span-based constituency parsing with a structure-label system and
  provably optimal dynamic oracles.
\newblock In \emph{EMNLP}.

\bibitem[{Devlin et~al.(2019)Devlin, Chang, Lee, and
  Toutanova}]{devlin2019bert}
Jacob Devlin, Ming-Wei Chang, Kenton Lee, and Kristina Toutanova. 2019.
\newblock Bert: Pre-training of deep bidirectional transformers for language
  understanding.
\newblock In \emph{NAACL}.

\bibitem[{Dominey and Dodane(2004)}]{DOMINEY2004121}
Peter~F Dominey and Christelle Dodane. 2004.
\newblock Indeterminacy in language acquisition: the role of child directed
  speech and joint attention.
\newblock \emph{Journal of Neurolinguistics}, 17(2):121--145.

\bibitem[{Drozdov et~al.(2019)Drozdov, Verga, Yadav, Iyyer, and
  McCallum}]{drozdov2019unsupervised}
Andrew Drozdov, Patrick Verga, Mohit Yadav, Mohit Iyyer, and Andrew McCallum.
  2019.
\newblock Unsupervised latent tree induction with deep inside-outside recursive
  auto-encoders.
\newblock In \emph{NAACL}.

\bibitem[{Gabeur et~al.(2020)Gabeur, Sun, Alahari, and
  Schmid}]{gabeur2020multi}
Valentin Gabeur, Chen Sun, Karteek Alahari, and Cordelia Schmid. 2020.
\newblock Multi-modal transformer for video retrieval.
\newblock In \emph{ECCV}.

\bibitem[{Gentner et~al.(2001)Gentner, Boroditsky, Bowerman, and
  Levinson}]{gentner2001individuation}
Dedre Gentner, Lera Boroditsky, Melissa Bowerman, and Stephen Levinson. 2001.
\newblock Individuation, relativity, and early word.
\newblock \emph{Language, culture and cognition}, 3:215--256.

\bibitem[{Glorot and Bengio(2010)}]{glorot2010understanding}
Xavier Glorot and Yoshua Bengio. 2010.
\newblock Understanding the difficulty of training deep feedforward neural
  networks.
\newblock In \emph{AISTATS}.

\bibitem[{Hendricks et~al.(2017)Hendricks, Wang, Shechtman, Sivic, Darrell, and
  Russell}]{hendricks17iccv}
Lisa~Anne Hendricks, Oliver Wang, Eli Shechtman, Josef Sivic, Trevor Darrell,
  and Bryan Russell. 2017.
\newblock Localizing moments in video with natural language.
\newblock In \emph{ICCV}.

\bibitem[{Hu et~al.(2018)Hu, Shen, and Sun}]{hu2018senet}
Jie Hu, Li~Shen, and Gang Sun. 2018.
\newblock Squeeze-and-excitation networks.
\newblock In \emph{CVPR}.

\bibitem[{Huang et~al.(2017)Huang, Liu, Van Der~Maaten, and
  Weinberger}]{huang2017densely}
Gao Huang, Zhuang Liu, Laurens Van Der~Maaten, and Kilian~Q Weinberger. 2017.
\newblock Densely connected convolutional networks.
\newblock In \emph{CVPR}.

\bibitem[{Huang et~al.(2021)Huang, Niu, Liu, Ding, Xiao, Wu, and
  Peng}]{NEURIPS2021_f5e62af8}
Zhenyu Huang, Guocheng Niu, Xiao Liu, Wenbiao Ding, Xinyan Xiao, Hua Wu, and
  Xi~Peng. 2021.
\newblock Learning with noisy correspondence for cross-modal matching.
\newblock In \emph{NeurIPS}.

\bibitem[{Jiao et~al.(2020)Jiao, Yin, Shang, Jiang, Chen, Li, Wang, and
  Liu}]{jiao2020tinybert}
Xiaoqi Jiao, Yichun Yin, Lifeng Shang, Xin Jiang, Xiao Chen, Linlin Li, Fang
  Wang, and Qun Liu. 2020.
\newblock Tinybert: Distilling bert for natural language understanding.
\newblock In \emph{EMNLP: Findings}.

\bibitem[{Jin et~al.(2018)Jin, Doshi-Velez, Miller, Schuler, and
  Schwartz}]{jin2018unsupervised}
Lifeng Jin, Finale Doshi-Velez, Timothy Miller, William Schuler, and Lane
  Schwartz. 2018.
\newblock Unsupervised grammar induction with depth-bounded {PCFG}.
\newblock \emph{TACL}.

\bibitem[{Jin et~al.(2019)Jin, Doshi-Velez, Miller, Schwartz, and
  Schuler}]{Jin2019-us}
Lifeng Jin, Finale Doshi-Velez, Timothy Miller, Lane Schwartz, and William
  Schuler. 2019.
\newblock Unsupervised learning of {PCFGs} with normalizing flow.
\newblock In \emph{{ACL}}.

\bibitem[{Jin and Schuler(2020)}]{jin2020grounded}
Lifeng Jin and William Schuler. 2020.
\newblock Grounded {PCFG} induction with images.
\newblock In \emph{AACL-IJCNLP}.

\bibitem[{Johnson et~al.(2007)Johnson, Griffiths, and
  Goldwater}]{johnson2007bayesian}
Mark Johnson, Thomas~L Griffiths, and Sharon Goldwater. 2007.
\newblock Bayesian inference for {PCFG}s via markov chain monte carlo.
\newblock In \emph{NAACL}.

\bibitem[{Kasami(1966)}]{kasami1966efficient}
Tadao Kasami. 1966.
\newblock An efficient recognition and syntax-analysis algorithm for
  context-free languages.
\newblock \emph{Coordinated Science Laboratory Report no. R-257}.

\bibitem[{Kim et~al.(2019{\natexlab{a}})Kim, Dyer, and Rush}]{kim2019compound}
Yoon Kim, Chris Dyer, and Alexander~M Rush. 2019{\natexlab{a}}.
\newblock Compound probabilistic context-free grammars for grammar induction.
\newblock In \emph{ACL}.

\bibitem[{Kim et~al.(2019{\natexlab{b}})Kim, Rush, Yu, Kuncoro, Dyer, and
  Melis}]{kim2019unsupervised}
Yoon Kim, Alexander~M Rush, Lei Yu, Adhiguna Kuncoro, Chris Dyer, and G{\'a}bor
  Melis. 2019{\natexlab{b}}.
\newblock Unsupervised recurrent neural network grammars.
\newblock In \emph{NAACL}.

\bibitem[{Kingma and Ba(2015)}]{kingma2014adam}
Diederik~P Kingma and Jimmy Ba. 2015.
\newblock Adam: A method for stochastic optimization.
\newblock In \emph{ICLR}.

\bibitem[{Kingma and Welling(2014)}]{kingma2013auto}
Diederik~P Kingma and Max Welling. 2014.
\newblock Auto-encoding variational bayes.
\newblock In \emph{ICLR}.

\bibitem[{Kitaev et~al.(2019)Kitaev, Cao, and
  Klein}]{kitaev-etal-2019-multilingual}
Nikita Kitaev, Steven Cao, and Dan Klein. 2019.
\newblock Multilingual constituency parsing with self-attention and
  pre-training.
\newblock In \emph{ACL}.

\bibitem[{Kitaev and Klein(2018)}]{kitaev2018constituency}
Nikita Kitaev and Dan Klein. 2018.
\newblock Constituency parsing with a self-attentive encoder.
\newblock In \emph{ACL}.

\bibitem[{Klein and Manning(2002)}]{klein2002generative}
Dan Klein and Christopher~D Manning. 2002.
\newblock A generative constituent-context model for improved grammar
  induction.
\newblock In \emph{ACL}.

\bibitem[{Kurihara and Sato(2006)}]{kurihara2006variational}
Kenichi Kurihara and Taisuke Sato. 2006.
\newblock Variational bayesian grammar induction for natural language.
\newblock In \emph{ICGI}.

\bibitem[{Lari and Young(1990)}]{lari1990estimation}
Karim Lari and Steve~J Young. 1990.
\newblock The estimation of stochastic context-free grammars using the
  inside-outside algorithm.
\newblock \emph{Computer speech \& language}, 4(1):35--56.

\bibitem[{Lin et~al.(2014)Lin, Maire, Belongie, Hays, Perona, Ramanan,
  Doll{\'a}r, and Zitnick}]{lin2014microsoft}
Tsung-Yi Lin, Michael Maire, Serge~J Belongie, James Hays, Pietro Perona, Deva
  Ramanan, Piotr Doll{\'a}r, and C~Lawrence Zitnick. 2014.
\newblock Microsoft coco: Common objects in context.
\newblock In \emph{ECCV}.

\bibitem[{Marcus et~al.(1993)Marcus, Santorini, and
  Marcinkiewicz}]{marcus1993building}
Mitch Marcus, Beatrice Santorini, and Mary~Ann Marcinkiewicz. 1993.
\newblock Building a large annotated corpus of english: The penn treebank.
\newblock \emph{Computational Linguistics}, 19(2):313--330.

\bibitem[{Miech et~al.(2020)Miech, Alayrac, Smaira, Laptev, Sivic, and
  Zisserman}]{miech2020end}
Antoine Miech, Jean-Baptiste Alayrac, Lucas Smaira, Ivan Laptev, Josef Sivic,
  and Andrew Zisserman. 2020.
\newblock End-to-end learning of visual representations from uncurated
  instructional videos.
\newblock In \emph{CVPR}.

\bibitem[{Miech et~al.(2018)Miech, Laptev, and Sivic}]{miech2018learning}
Antoine Miech, Ivan Laptev, and Josef Sivic. 2018.
\newblock Learning a text-video embedding from incomplete and heterogeneous
  data.
\newblock \emph{arXiv preprint arXiv:1804.02516}.

\bibitem[{Miech et~al.(2019)Miech, Zhukov, Alayrac, Tapaswi, Laptev, and
  Sivic}]{miech2019howto100m}
Antoine Miech, Dimitri Zhukov, Jean-Baptiste Alayrac, Makarand Tapaswi, Ivan
  Laptev, and Josef Sivic. 2019.
\newblock {HowTo100M}: Learning a text-video embedding by watching hundred
  million narrated video clips.
\newblock In \emph{ICCV}.

\bibitem[{Petrov and Klein(2007)}]{petrov2007improved}
Slav Petrov and Dan Klein. 2007.
\newblock Improved inference for unlexicalized parsing.
\newblock In \emph{NAACL}.

\bibitem[{Radford et~al.(2021)Radford, Kim, Hallacy, Ramesh, Goh, Agarwal,
  Sastry, Askell, Mishkin, Clark et~al.}]{radford2021learning}
Alec Radford, Jong~Wook Kim, Chris Hallacy, Aditya Ramesh, Gabriel Goh,
  Sandhini Agarwal, Girish Sastry, Amanda Askell, Pamela Mishkin, Jack Clark,
  et~al. 2021.
\newblock Learning transferable visual models from natural language
  supervision.
\newblock \emph{arXiv preprint arXiv:2103.00020}.

\bibitem[{Schroff et~al.(2015)Schroff, Kalenichenko, and
  Philbin}]{schroff2015facenet}
Florian Schroff, Dmitry Kalenichenko, and James Philbin. 2015.
\newblock Facenet: A unified embedding for face recognition and clustering.
\newblock In \emph{CVPR}.

\bibitem[{Shen et~al.(2018{\natexlab{a}})Shen, Lin, wei Huang, and
  Courville}]{shen2018neural}
Yikang Shen, Zhouhan Lin, Chin wei Huang, and Aaron Courville.
  2018{\natexlab{a}}.
\newblock Neural language modeling by jointly learning syntax and lexicon.
\newblock In \emph{ICLR}.

\bibitem[{Shen et~al.(2018{\natexlab{b}})Shen, Tan, Sordoni, and
  Courville}]{shen2018ordered}
Yikang Shen, Shawn Tan, Alessandro Sordoni, and Aaron Courville.
  2018{\natexlab{b}}.
\newblock Ordered neurons: Integrating tree structures into recurrent neural
  networks.
\newblock In \emph{ICLR}.

\bibitem[{Shi et~al.(2020)Shi, Livescu, and Gimpel}]{shi2020role}
Haoyue Shi, Karen Livescu, and Kevin Gimpel. 2020.
\newblock On the role of supervision in unsupervised constituency parsing.
\newblock In \emph{EMNLP}.

\bibitem[{Shi et~al.(2019)Shi, Mao, Gimpel, and Livescu}]{shi2019visually}
Haoyue Shi, Jiayuan Mao, Kevin Gimpel, and Karen Livescu. 2019.
\newblock Visually grounded neural syntax acquisition.
\newblock In \emph{ACL}.

\bibitem[{Tilk and Alum{\"a}e(2016)}]{tilk2016}
Ottokar Tilk and Tanel Alum{\"a}e. 2016.
\newblock Bidirectional recurrent neural network with attention mechanism for
  punctuation restoration.
\newblock In \emph{Interspeech}.

\bibitem[{Tran et~al.(2018)Tran, Wang, Torresani, Ray, LeCun, and
  Paluri}]{tran2018closer}
Du~Tran, Heng Wang, Lorenzo Torresani, Jamie Ray, Yann LeCun, and Manohar
  Paluri. 2018.
\newblock A closer look at spatiotemporal convolutions for action recognition.
\newblock In \emph{CVPR}.

\bibitem[{Vaswani et~al.(2017)Vaswani, Shazeer, Parmar, Uszkoreit, Jones,
  Gomez, Kaiser, and Polosukhin}]{vaswani2017attention}
Ashish Vaswani, Noam Shazeer, Niki Parmar, Jakob Uszkoreit, Llion Jones,
  Aidan~N Gomez, {\L}ukasz Kaiser, and Illia Polosukhin. 2017.
\newblock Attention is all you need.
\newblock In \emph{NIPS}.

\bibitem[{Wang and Blunsom(2013)}]{wang2013collapsed}
Pengyu Wang and Phil Blunsom. 2013.
\newblock Collapsed variational bayesian inference for {PCFG}s.
\newblock In \emph{CoNLL}.

\bibitem[{Wang et~al.(2019)Wang, Xie, Song, Zang, Wang, Lu, Yu, and
  Shen}]{wang2019efficient}
Wenhai Wang, Enze Xie, Xiaoge Song, Yuhang Zang, Wenjia Wang, Tong Lu, Gang Yu,
  and Chunhua Shen. 2019.
\newblock Efficient and accurate arbitrary-shaped text detection with pixel
  aggregation network.
\newblock In \emph{ICCV}.

\bibitem[{Xie et~al.(2017)Xie, Girshick, Doll{\'a}r, Tu, and
  He}]{xie2017aggregated}
Saining Xie, Ross Girshick, Piotr Doll{\'a}r, Zhuowen Tu, and Kaiming He. 2017.
\newblock Aggregated residual transformations for deep neural networks.
\newblock In \emph{CVPR}.

\bibitem[{Xu et~al.(2016)Xu, Mei, Yao, and Rui}]{xu2016msr}
Jun Xu, Tao Mei, Ting Yao, and Yong Rui. 2016.
\newblock {MSR-VTT}: A large video description dataset for bridging video and
  language.
\newblock In \emph{CVPR}.

\bibitem[{Yang et~al.(2021{\natexlab{a}})Yang, Zhao, and
  Tu}]{yang-etal-2021-neural}
Songlin Yang, Yanpeng Zhao, and Kewei Tu. 2021{\natexlab{a}}.
\newblock Neural bi-lexicalized {PCFG} induction.
\newblock In \emph{ACL}.

\bibitem[{Yang et~al.(2021{\natexlab{b}})Yang, Zhao, and
  Tu}]{yang-etal-2021-pcfgs}
Songlin Yang, Yanpeng Zhao, and Kewei Tu. 2021{\natexlab{b}}.
\newblock {PCFG}s can do better: Inducing probabilistic context-free grammars
  with many symbols.
\newblock In \emph{NAACL}.

\bibitem[{Younger(1967)}]{younger1967recognition}
Daniel~H Younger. 1967.
\newblock Recognition and parsing of context-free languages in time n3.
\newblock \emph{Information and control}, 10(2):189--208.

\bibitem[{Yu and Siskind(2013)}]{yu2013grounded}
Haonan Yu and Jeffrey~Mark Siskind. 2013.
\newblock Grounded language learning from video described with sentences.
\newblock In \emph{ACL}.

\bibitem[{Zhang et~al.(2017)Zhang, Zhang, Wang, Li, Qiao, and
  Liu}]{zhang2017detecting}
Kaipeng Zhang, Zhanpeng Zhang, Hao Wang, Zhifeng Li, Yu~Qiao, and Wei Liu.
  2017.
\newblock Detecting faces using inside cascaded contextual cnn.
\newblock In \emph{ICCV}.

\bibitem[{Zhang et~al.(2021)Zhang, Song, Jin, Xu, Yu, and Luo}]{zhang2021video}
Songyang Zhang, Linfeng Song, Lifeng Jin, Kun Xu, Dong Yu, and Jiebo Luo. 2021.
\newblock Video-aided unsupervised grammar induction.
\newblock In \emph{NAACL}.

\bibitem[{Zhang et~al.(2019)Zhang, Su, and Luo}]{zhang2019exploiting}
Songyang Zhang, Jinsong Su, and Jiebo Luo. 2019.
\newblock Exploiting temporal relationships in video moment localization with
  natural language.
\newblock In \emph{ACMMM}.

\bibitem[{Zhang and Clark(2011)}]{zhang2011syntactic}
Yue Zhang and Stephen Clark. 2011.
\newblock Syntactic processing using the generalized perceptron and beam
  search.
\newblock \emph{Computational linguistics}, 37(1):105--151.

\bibitem[{Zhao and Titov(2020)}]{zhao2020visually}
Yanpeng Zhao and Ivan Titov. 2020.
\newblock Visually grounded compound {PCFG}s.
\newblock In \emph{EMNLP}.

\bibitem[{Zhou et~al.(2018)Zhou, Xu, and Corso}]{ZhXuCoAAAI18}
Luowei Zhou, Chenliang Xu, and Jason~J Corso. 2018.
\newblock Towards automatic learning of procedures from web instructional
  videos.
\newblock In \emph{AAAI}.

\end{thebibliography}
\bibliographystyle{acl_natbib}

\newpage
\appendix

\onecolumn


\section{Video Processing Details}
\label{appendix:preprocessing_details}
\noindent\textbf{MIL-NCE.} \revision{
Following the implementation\footnote{\url{https://github.com/antoine77340/S3D\_HowTo100M}} of MIL-NCE, we extract $1$ feature per second from video encoder's last fully connected layer. All videos are decoded at $16$ frames per second (fps).
}

\noindent\textbf{MM.} \revision{
We list the details of object, action, scene, OCR and face feature extraction as below:
\begin{itemize}
    \item ResNeXt~\cite{xie2017aggregated}. We use the \textit{ResNeXt101} version implemented by torchvision\footnote{\url{https://github.com/pytorch/vision}}. Videos are decoded at $1$ fps and we extract $1$ feature per second from the last fully connected layer. 
    \item SENet~\cite{hu2018senet}. We use the \textit{SENet154} version implemented by Cadene\footnote{\url{https://github.com/Cadene/pretrained-models.pytorch}}. Videos are decoded at $1$ fps and we extract $1$ feature per second from the last fully connected layer. 
    \item I3D\cite{carreira2017quo}. We use the \textit{I3D\_8x8\_R50} version implemented by SlowFast\footnote{\url{https://github.com/facebookresearch/SlowFast}}. We decode videos at $4$ fps and extract $1$ feature per $2$ seconds from the last fully connected layer. 
    \item S3DG~\cite{miech2020end}. We use the implementation from HERO\footnote{\url{https://github.com/linjieli222/HERO_Video_Feature_Extractor}}. Videos are decoded at $30$ fps and we extract $1$ feature per $1.5$ seconds from the global averaged pooling layer. 
    \item R2P1D~\cite{tran2018closer}. We use the \textit{r2plus1d\_34} version implemented by torchvision. we decode videos at $16$ fps and extract $1$ feature per $2$ seconds. 
    \item Scene. We use \textit{densenet161}~\cite{huang2017densely} implemented by CSAILVision\footnote{\url{https://github.com/CSAILVision/places365}}. Videos are decoded at $1$ fps and we extract $1$ feature per second from the last fully connected layer. 
    \item OCR. We use the text detector PANet~\cite{wang2019efficient} and the text recognizer \textit{seg\_r31} implemented by MMOCR\footnote{\url{https://github.com/open-mmlab/mmocr}}. we decode videos at $0.5$ fps and extract $1$ feature per $2$ seconds.
    \item Face. We use face detector MTCNN~\cite{zhang2017detecting} and face recognizer~\cite{schroff2015facenet} implemented by FaceNet\footnote{\url{https://github.com/timesler/facenet-pytorch}}. We decode videos at $1$ fps and extract $1$ feature per second.
\end{itemize}
}

\noindent\textbf{CLIP.} \revision{Following the implementation\footnote{\url{https://github.com/openai/CLIP}} of CLIP, we extract $1$ video feature per second from \textit{ViT-B/32}'s last fully connected layer. All videos are decoded at $1$ fps.}

\section{Computational Cost}
\label{appendix:computational_cost}
All our models are trained on $2$ $32$GB V$100$ GPUs. 
The approximate time cost for each run of different model is listed in Table~\ref{tab:time}. For each model, we run \revision{$5$} times with different random seeds in parallel. During training, the video encoder, the span encoder and C-PCFG are involved, which contains $76.6$M parameters in total. During inference, since only C-PCFG is involved, there are $23.0$M parameters in total.

\setcounter{table}{0}
\begin{table}[h]
\small
    \centering
    \caption{The approximate training time (hours) of different model on a single run.}
    \begin{tabular}{cccccc}
    \toprule
         Model &  HT(29.6k) & HT(296k) & HT(592k) & HT(1.2M) & HT(2.4M)\\
     \midrule
         C-PCFG & $0.07$ & $0.7$ & $1.5$ & $2.9$ & $5.8$ \\
         MMC-PCFG & $0.75$ & $7.5$ & $15$ & $30$ & $60$ \\
         PTC-PCFG & $0.50$ & $5.0$ & $10$ & $20$ & $40$ \\
     \bottomrule
    \end{tabular}
    \label{tab:time}
\end{table}



\section{Performance Comparison - Full Tables}
\label{appendix:full_table}
We compare the performances of different models trained on different datasets. The full experiment results are demonstrated in Table~\ref{tab:DiDeMo}-\ref{tab:MSRVTT}. LBranch, RBranch and Random represent left branching tree, right branching tree and random tree, respectively. In addition to C-F1 and S-F1, we also evaluate the recall of each model on different phrase types, including NP, VP, PP, SBAR, ADJP and ADVP. All numbers are shown in percentage($\%$).

\begin{table*}[hbt!]
\small
\setlength{\tabcolsep}{4pt}
    \centering
    \caption{Performance comparison on DiDeMo.}
    \begin{tabular}{ccccccccccc}
    \toprule
	\multicolumn{2}{c}{Method} & Trainset & NP & VP & PP & SBAR & ADJP & ADVP & C-F1 & S-F1 \\
	\midrule
    \multicolumn{2}{c}{LBranch} & None & $41.7$ & $0.1 $ & $0.1 $ & $0.7 $ & $7.2 $ & $0.0 $ & $16.2$ & $18.5$ \\
    \multicolumn{2}{c}{RBranch} & None & $32.8$ & $\mathbf{91.5}$ & ${66.5}$ & $\mathbf{88.2}$ & ${36.9}$ & $\mathbf{63.6}$ & ${53.6}$ & ${57.5}$ \\
    \multicolumn{2}{c}{Random} & None  & $36.5_{\pm0.6 }$ & $30.5_{\pm0.5 }$ & $30.1_{\pm0.5 }$ & $25.7_{\pm2.8 }$ & $29.5_{\pm2.3 }$ & $28.5_{\pm4.8 }$ & $29.4_{\pm0.3 }$ & $32.7_{\pm0.5 }$ \\
    \multicolumn{2}{c}{C-PCFG} & DiDeMo & ${72.9}_{\pm5.5 }$ & $16.5_{\pm6.2 }$ & $23.4_{\pm16.9}$ & $26.6_{\pm15.9}$ & $25.0_{\pm11.6}$ & $14.7_{\pm12.8}$ & $38.2_{\pm5.0 }$ & $40.4_{\pm4.1 }$ \\
    \multirow{11}{*}{\rotatebox{90}{VC-PCFG}} & ResNeXt & DiDeMo & $64.4_{\pm21.4}$ & $25.7_{\pm17.7}$ & $34.6_{\pm25.0}$ & $40.5_{\pm26.3}$ & $16.7_{\pm9.5 }$ & $28.4_{\pm21.3}$ & $40.0_{\pm13.7}$ & $41.8_{\pm14.0}$ \\
    & SENet & DiDeMo & $70.5_{\pm15.3}$ & $25.7_{\pm15.9}$ & $36.5_{\pm24.6}$ & $36.8_{\pm25.9}$ & $21.2_{\pm12.5}$ & $23.6_{\pm16.8}$ & $42.6_{\pm10.4}$ & $44.0_{\pm10.4}$ \\
    & I3D & DiDeMo & $57.9_{\pm13.5}$ & $45.7_{\pm14.1}$ & $45.8_{\pm17.2}$ & $38.2_{\pm14.8}$ & $28.4_{\pm9.2 }$ & $22.0_{\pm9.3 }$ & $45.1_{\pm6.0 }$ & $49.2_{\pm6.0 }$ \\
    & R2P1D & DiDeMo & $61.2_{\pm8.5 }$ & $38.1_{\pm5.4 }$ & $62.1_{\pm4.1 }$ & $61.5_{\pm5.1 }$ & $21.4_{\pm11.4}$ & $40.8_{\pm7.3 }$ & $48.1_{\pm4.4 }$ & $50.7_{\pm4.2 }$ \\
    & S3DG & DiDeMo & $61.3_{\pm13.4}$ & $31.7_{\pm16.7}$ & $51.8_{\pm8.0 }$ & $50.3_{\pm6.5 }$ & $18.0_{\pm4.5 }$ & $35.2_{\pm11.4}$ & $44.0_{\pm2.7 }$ & $46.5_{\pm5.1 }$ \\
    & Scene & DiDeMo & $62.2_{\pm9.6 }$ & $30.6_{\pm12.3}$ & $41.1_{\pm24.8}$ & $35.2_{\pm21.9}$ & $21.4_{\pm14.0}$ & $27.6_{\pm17.1}$ & $41.7_{\pm6.5 }$ & $44.9_{\pm7.4 }$ \\
    & Audio & DiDeMo & $64.2_{\pm18.6}$ & $21.3_{\pm26.5}$ & $34.7_{\pm11.0}$ & $37.3_{\pm19.6}$ & $26.1_{\pm4.9 }$ & $18.2_{\pm11.6}$ & $38.7_{\pm3.7 }$ & $39.5_{\pm5.2 }$ \\
    & OCR & DiDeMo & $64.4_{\pm15.0}$ & $27.4_{\pm19.5}$ & $42.8_{\pm31.2}$ & $35.9_{\pm20.7}$ & $14.6_{\pm1.7 }$ & $23.2_{\pm24.0}$ & $41.9_{\pm16.9}$ & $44.6_{\pm17.5}$ \\
    & Face & DiDeMo & $60.8_{\pm16.0}$ & $31.5_{\pm17.0}$ & $52.8_{\pm9.8 }$ & $49.3_{\pm5.6 }$ & $12.6_{\pm3.3 }$ & $32.9_{\pm14.6}$ & $43.9_{\pm4.5 }$ & $46.3_{\pm5.5 }$ \\
    & Speech & DiDeMo & $61.8_{\pm12.8}$ & $26.6_{\pm17.6}$ & $43.8_{\pm34.5}$ & $34.2_{\pm20.6}$ & $14.4_{\pm4.8 }$ & $12.9_{\pm9.6 }$ & $40.9_{\pm16.0}$ & $43.1_{\pm16.1}$ \\
    & Concat & DiDeMo & $68.6_{\pm8.6 }$ & $24.9_{\pm19.9}$ & $39.7_{\pm19.5}$ & $39.3_{\pm19.8}$ & $10.8_{\pm2.8 }$ & $18.3_{\pm18.1}$ & $42.2_{\pm12.3}$ & $43.2_{\pm14.2}$ \\
    \multicolumn{2}{c}{MMC-PCFG} & DiDeMo & ${67.9}_{\pm9.8 }$ & ${52.3}_{\pm9.0 }$ & ${63.5}_{\pm8.6 }$ & ${60.7}_{\pm10.8}$ & ${34.7}_{\pm17.0}$ & ${50.4}_{\pm8.3 }$ & ${55.0}_{\pm3.7 }$ & ${58.9}_{\pm3.4 }$ \\
    \multicolumn{2}{c}{MMC-PCFG} & YouCook2 & $47.9_{\pm10.4}$ & $34.6_{\pm2.7}$ & $58.2_{\pm12.9}$ & $19.9_{\pm2.6}$ & $11.0_{\pm4.0}$ & $25.5_{\pm4.8}$ & $40.1_{\pm4.4}$ & $44.2_{\pm4.4}$ \\
    \multicolumn{2}{c}{MMC-PCFG} & MSRVTT & $56.5_{\pm6.8}$ & $\mathit{70.8}_{\pm4.4}$ & $\mathbf{82.6}_{\pm3.3}$ & $\mathit{62.5}_{\pm6.0}$ & $42.6_{\pm2.6}$ & $52.9_{\pm7.6}$ & $59.4_{\pm2.9}$ & $62.7_{\pm3.3}$ \\ 
    \midrule

    \multicolumn{2}{c}{C-PCFG} & HT(29.6k) & $74.8_{\pm1.1}$ & $27.3_{\pm5.7}$ & $43.6_{\pm13.2}$ & $32.9_{\pm7.4}$ & $32.4_{\pm4.6}$ & $44.5_{\pm7.8}$ & $45.7_{\pm3.9}$ & $47.7_{\pm3.5}$ \\
    \multicolumn{2}{c}{MMC-PCFG} & HT(29.6k) & $75.4_{\pm2.1}$ & $33.2_{\pm12.5}$ & $54.9_{\pm7.8}$ & $33.7_{\pm9.8}$ & $39.2_{\pm4.7}$ & $43.8_{\pm7.1}$ & $49.8_{\pm4.7}$ & $52.3_{\pm5.8}$ \\
    \multicolumn{2}{c}{\textbf{PTC-PCFG}} & HT(29.6k) & $66.0_{\pm9.4}$ & $53.7_{\pm13.9}$ & $68.2_{\pm4.5}$ & $50.4_{\pm5.0}$ & $35.2_{\pm4.2}$ & $52.7_{\pm8.9}$ & $54.9_{\pm3.4}$ & $58.5_{\pm4.0}$ \\
    \midrule

    \multicolumn{2}{c}{C-PCFG} & HT(296k) & $81.4_{\pm1.6}$ & $36.4_{\pm6.9}$ & $67.0_{\pm1.9}$ & $45.9_{\pm3.5}$ & $46.5_{\pm4.8}$ & $49.9_{\pm8.3}$ & $55.6_{\pm1.4}$ & $58.5_{\pm1.9}$ \\
    \multicolumn{2}{c}{MMC-PCFG} & HT(296k) & $81.9_{\pm2.1}$ & $42.7_{\pm15.0}$ & $65.3_{\pm6.4}$ & $39.1_{\pm8.5}$ & $\mathit{48.0}_{\pm8.1}$ & $43.7_{\pm6.7}$ & $57.1_{\pm4.2}$ & $59.9_{\pm4.8}$ \\
    \multicolumn{2}{c}{\textbf{PTC-PCFG}} & HT(296k) & $76.0_{\pm4.9}$ & $55.3_{\pm11.9}$ & $70.7_{\pm6.4}$ & $53.7_{\pm9.5}$ & $43.4_{\pm4.8}$ & $47.2_{\pm13.3}$ & $59.5_{\pm4.3}$ & $63.7_{\pm4.7}$ \\
    \midrule

    \multicolumn{2}{c}{C-PCFG} & HT(592k) & $82.4_{\pm1.9}$ & $37.6_{\pm9.2}$ & $63.1_{\pm5.8}$ & $37.4_{\pm7.9}$ & $45.8_{\pm5.5}$ & $\mathit{53.8}_{\pm9.9}$ & $55.5_{\pm2.7}$ & $57.5_{\pm2.9}$ \\
    \multicolumn{2}{c}{MMC-PCFG} & HT(592k) & $79.0_{\pm5.7}$ & $52.5_{\pm19.1}$ & $64.3_{\pm6.1}$ & $45.5_{\pm9.9}$ & $44.1_{\pm4.4}$ & $44.5_{\pm11.7}$ & $58.5_{\pm7.3}$ & $62.4_{\pm7.9}$ \\
    \multicolumn{2}{c}{\textbf{PTC-PCFG}} & HT(592k) & $79.1_{\pm2.9}$ & $55.3_{\pm18.4}$ & $73.7_{\pm5.1}$ & $50.6_{\pm8.2}$ & $38.3_{\pm5.6}$ & $47.2_{\pm5.5}$ & $\mathbf{61.3}_{\pm3.9}$ & $\mathbf{65.2}_{\pm5.3}$ \\
    \midrule

    \multicolumn{2}{c}{C-PCFG} & HT(1.2M) & $\mathit{82.9}_{\pm1.5}$ & $32.2_{\pm12.5}$ & $67.5_{\pm2.1}$ & $41.3_{\pm8.5}$ & $45.6_{\pm4.8}$ & $48.8_{\pm8.0}$ & $55.1_{\pm3.3}$ & $57.2_{\pm3.9}$ \\
    \multicolumn{2}{c}{MMC-PCFG} & HT(1.2M) & $\mathbf{83.4}_{\pm2.6}$ & $43.2_{\pm15.9}$ & $51.3_{\pm19.1}$ & $36.2_{\pm7.0}$ & $\mathbf{49.4}_{\pm5.5}$ & $56.2_{\pm6.9}$ & $55.1_{\pm5.0}$ & $58.2_{\pm5.8}$ \\
    \multicolumn{2}{c}{\textbf{PTC-PCFG}} & HT(1.2M) & $77.9_{\pm1.9}$ & $49.0_{\pm7.6}$ & $\mathit{78.3}_{\pm3.9}$ & $47.2_{\pm6.5}$ & $35.8_{\pm8.7}$ & $49.8_{\pm11.6}$ & $\mathit{60.0}_{\pm2.3}$ & $\mathit{64.2}_{\pm3.1}$ \\
    \midrule

    \multicolumn{2}{c}{C-PCFG} & HT(2.4M) & $\mathbf{83.4}_{\pm2.2}$ & $31.0_{\pm4.7}$ & $68.1_{\pm8.4}$ & $48.5_{\pm7.7}$ & $46.9_{\pm7.3}$ & $45.9_{\pm8.2}$ & $55.2_{\pm3.0}$ & $56.8_{\pm3.8}$ \\
    \multicolumn{2}{c}{MMC-PCFG} & HT(2.4M) & $81.7_{\pm2.8}$ & $37.6_{\pm4.8}$ & $71.1_{\pm6.9}$ & $47.2_{\pm6.1}$ & $41.7_{\pm8.1}$ & $51.2_{\pm7.2}$ & $57.0_{\pm2.1}$ & $58.6_{\pm2.1}$ \\
    \multicolumn{2}{c}{\textbf{PTC-PCFG}} & HT(2.4M) & $78.9_{\pm1.6}$ & $49.4_{\pm6.9}$ & $75.0_{\pm5.2}$ & $48.7_{\pm6.0}$ & $36.9_{\pm5.4}$ & $51.8_{\pm13.8}$ & $\mathit{60.0}_{\pm2.5}$ & $63.1_{\pm2.3}$ \\
    \bottomrule
    
    \end{tabular}
    \label{tab:DiDeMo}
\end{table*}

\begin{table*}[hbt!]
\small
\setlength{\tabcolsep}{4pt}
    \centering
    \caption{Performance comparison on YouCook2.}
    \begin{tabular}{ccccccccccc}
    \toprule
	\multicolumn{2}{c}{Method} & Trainset & NP & VP & PP & SBAR & ADJP & ADVP & C-F1 & S-F1 \\
	\hline
    \multicolumn{2}{c}{LBranch} & None & $1.7 $ & $42.8$ & $0.4 $ & $8.1 $ & $1.5 $ & $0.0 $ & $6.8 $ & $5.9 $ \\
    \multicolumn{2}{c}{RBranch} & None & $35.6$ & ${47.5}$ & $67.0$ & $\mathbf{88.9}$ & ${33.9}$ & ${65.0}$ & $35.0$ & $41.6$ \\
    \multicolumn{2}{c}{Random} & None & $27.2_{\pm0.3 }$ & $27.1_{\pm1.4 }$ & $29.9_{\pm0.5 }$ & $31.3_{\pm5.2 }$ & $26.9_{\pm7.7 }$ & $26.2_{\pm11.9}$ & $21.2_{\pm0.2}$ & $24.0_{\pm0.2}$ \\
    \multicolumn{2}{c}{C-PCFG} & YouCook2 & $47.4_{\pm18.4}$ & ${49.4}_{\pm11.9}$ & $58.0_{\pm22.6}$ & $45.7_{\pm6.0 }$ & ${27.7}_{\pm15.1}$ & $36.2_{\pm7.4 }$ & $37.8_{\pm6.7}$ & $41.4_{\pm6.6}$ \\
    \multirow{8}{*}{\rotatebox{90}{VC-PCFG}} & ResNeXt & YouCook2 & $46.5_{\pm13.7}$ & $40.8_{\pm9.8 }$ & $67.9_{\pm12.7}$ & $50.5_{\pm13.3}$ & $22.3_{\pm6.7 }$ & $38.8_{\pm21.3}$ & $38.2_{\pm8.3}$ & $42.8_{\pm8.4}$ \\
    & SENet & YouCook2 & $48.3_{\pm14.4}$ & $40.7_{\pm9.2 }$ & $73.6_{\pm11.2}$ & $45.5_{\pm17.0}$ & $26.9_{\pm13.6}$ & $41.2_{\pm17.5}$ & $39.9_{\pm8.7}$ & $44.9_{\pm8.3}$ \\
    & I3D & YouCook2 & $48.1_{\pm10.7}$ & $39.0_{\pm8.0 }$ & ${79.4}_{\pm8.4 }$ & $50.0_{\pm14.9}$ & $18.5_{\pm7.0 }$ & $41.2_{\pm4.1 }$ & $40.6_{\pm3.6}$ & $45.7_{\pm3.2}$ \\
    & R2P1D & YouCook2 & ${52.4}_{\pm10.9}$ & $33.7_{\pm16.4}$ & $66.7_{\pm10.7}$ & $49.5_{\pm13.8}$ & $25.8_{\pm10.6}$ & $33.8_{\pm12.4}$ & $39.4_{\pm8.1}$ & $44.4_{\pm8.3}$ \\
    & S3DG & YouCook2 & $50.4_{\pm13.1}$ & $32.6_{\pm16.3}$ & $71.7_{\pm7.5 }$ & $33.3_{\pm5.9 }$ & $30.8_{\pm17.5}$ & $40.0_{\pm7.1 }$ & $39.3_{\pm6.5}$ & $44.1_{\pm6.6}$ \\
    & Audio & YouCook2 & $51.2_{\pm3.1 }$ & $42.0_{\pm7.2 }$ & $61.5_{\pm18.0}$ & ${51.0}_{\pm14.8}$ & $23.5_{\pm16.8}$ & $48.8_{\pm8.2 }$ & $39.2_{\pm4.7}$ & $43.3_{\pm4.9}$ \\
    & OCR & YouCook2 & $48.6_{\pm8.1 }$ & $41.5_{\pm4.1 }$ & $65.5_{\pm17.4}$ & $39.9_{\pm4.4 }$ & $18.5_{\pm6.6 }$ & ${53.8}_{\pm14.7}$ & $38.6_{\pm5.5}$ & $43.2_{\pm5.6}$ \\
    & Concat & YouCook2 & $50.3_{\pm10.3}$ & $42.3_{\pm2.9 }$ & ${81.6}_{\pm8.7 }$ & $40.1_{\pm3.9 }$ & $17.7_{\pm8.2 }$ & $52.5_{\pm5.6 }$ & ${42.3}_{\pm5.7}$ & ${47.0}_{\pm5.6}$ \\
    \multicolumn{2}{c}{{MMC-PCFG}} & YouCook2 & ${62.7}_{\pm9.8 }$ & $45.3_{\pm2.8 }$ & $63.4_{\pm17.7}$ & $43.9_{\pm4.8 }$ & $26.2_{\pm7.5 }$ & $35.0_{\pm3.5 }$ & ${44.7}_{\pm5.2}$ & ${48.9}_{\pm5.7}$ \\
    \multicolumn{2}{c}{MMC-PCFG} & DiDeMo & $63.8_{\pm4.5}$ & $62.1_{\pm7.4}$ & $70.7_{\pm9.0}$ & $56.8_{\pm9.2}$ & $35.4_{\pm7.2}$ & $51.2_{\pm4.1}$ & $49.1_{\pm4.4}$ & $53.0_{\pm4.9}$ \\ 
    \multicolumn{2}{c}{MMC-PCFG} & MSRVTT & $63.1_{\pm9.2}$ & $51.5_{\pm7.3}$ & $82.7_{\pm1.5}$ & ${64.9}_{\pm10.8}$ & $30.4_{\pm3.0}$ & $40.0_{\pm6.1}$ & $49.6_{\pm3.9}$ & $54.2_{\pm4.1}$ \\ 
    \midrule

    \multicolumn{2}{c}{C-PCFG} & HT(29.6k) & $68.2_{\pm3.3}$ & $43.5_{\pm4.9}$ & $48.8_{\pm20.2}$ & $34.5_{\pm4.2}$ & $28.5_{\pm3.2}$ & $56.7_{\pm6.2}$ & $44.3_{\pm2.4}$ & $49.2_{\pm2.8}$ \\
    \multicolumn{2}{c}{MMC-PCFG} & HT(29.6k) & $68.8_{\pm3.0}$ & $51.3_{\pm11.5}$ & $65.5_{\pm11.2}$ & $33.7_{\pm6.3}$ & $32.0_{\pm4.3}$ & $55.0_{\pm11.3}$ & $48.8_{\pm3.5}$ & $53.6_{\pm3.6}$ \\
    \multicolumn{2}{c}{\textbf{PTC-PCFG}} & HT(29.6k) & $66.3_{\pm4.2}$ & $55.5_{\pm8.3}$ & $76.5_{\pm5.7}$ & $46.2_{\pm9.0}$ & $33.4_{\pm7.4}$ & $40.0_{\pm9.7}$ & $50.2_{\pm3.4}$ & $55.4_{\pm2.9}$ \\
    \midrule

    \multicolumn{2}{c}{C-PCFG} & HT(296k) & $77.5_{\pm3.1}$ & $56.6_{\pm4.5}$ & $74.4_{\pm5.5}$ & $51.1_{\pm9.0}$ & $40.5_{\pm6.4}$ & $63.3_{\pm8.5}$ & $55.0_{\pm2.7}$ & $60.5_{\pm2.5}$ \\
    \multicolumn{2}{c}{MMC-PCFG} & HT(296k) & $76.5_{\pm4.1}$ & $61.8_{\pm9.7}$ & $67.2_{\pm8.7}$ & $34.0_{\pm16.2}$ & $41.0_{\pm7.9}$ & $68.3_{\pm8.2}$ & $53.8_{\pm3.4}$ & $59.0_{\pm3.5}$ \\
    \multicolumn{2}{c}{\textbf{PTC-PCFG}} & HT(296k) & $77.5_{\pm2.5}$ & $65.8_{\pm6.1}$ & $78.9_{\pm7.0}$ & $61.6_{\pm3.9}$ & $42.4_{\pm6.3}$ & $60.0_{\pm3.3}$ & $57.1_{\pm1.7}$ & $62.1_{\pm1.3}$ \\
    \midrule

    \multicolumn{2}{c}{C-PCFG} & HT(592k) & $76.7_{\pm5.6}$ & $64.7_{\pm12.3}$ & $65.4_{\pm22.5}$ & $45.2_{\pm13.9}$ & $45.1_{\pm9.4}$ & $68.3_{\pm6.2}$ & $54.2_{\pm6.0}$ & $58.5_{\pm6.3}$ \\
    \multicolumn{2}{c}{MMC-PCFG} & HT(592k) & $72.5_{\pm12.7}$ & $68.9_{\pm7.0}$ & $68.3_{\pm18.2}$ & $54.3_{\pm5.7}$ & $\mathbf{50.2}_{\pm2.1}$ & $\mathit{75.0}_{\pm9.1}$ & $53.9_{\pm6.6}$ & $58.0_{\pm7.1}$ \\
    \multicolumn{2}{c}{\textbf{PTC-PCFG}} & HT(592k) & $78.7_{\pm5.3}$ & $69.9_{\pm3.6}$ & $80.5_{\pm2.8}$ & $58.9_{\pm12.3}$ & $43.2_{\pm4.0}$ & $65.0_{\pm6.2}$ & $58.9_{\pm2.5}$ & $63.2_{\pm2.3}$ \\
    \midrule

    \multicolumn{2}{c}{C-PCFG} & HT(1.2M) & $\mathit{80.1}_{\pm2.5}$ & $63.5_{\pm11.9}$ & $78.5_{\pm2.5}$ & $52.5_{\pm13.7}$ & $42.9_{\pm6.5}$ & $66.7_{\pm7.5}$ & $58.1_{\pm2.4}$ & $63.1_{\pm2.1}$ \\
    \multicolumn{2}{c}{MMC-PCFG} & HT(1.2M) & $75.7_{\pm2.2}$ & $64.8_{\pm9.7}$ & $57.7_{\pm21.4}$ & $51.2_{\pm9.4}$ & $\mathit{49.3}_{\pm4.9}$ & $\mathit{75.0}_{\pm7.5}$ & $52.5_{\pm4.0}$ & $57.4_{\pm4.3}$ \\
    \multicolumn{2}{c}{\textbf{PTC-PCFG}} & HT(1.2M) & $78.1_{\pm2.6}$ & $\mathbf{72.2}_{\pm4.1}$ & $\mathbf{85.8}_{\pm5.1}$ & $\mathit{69.3}_{\pm6.6}$ & $41.5_{\pm8.7}$ & $66.7_{\pm10.5}$ & $\mathit{60.1}_{\pm1.4}$ & $\mathit{64.5}_{\pm1.3}$ \\
    \midrule

    \multicolumn{2}{c}{C-PCFG} & HT(2.4M) & $75.9_{\pm3.9}$ & $61.5_{\pm8.0}$ & $78.2_{\pm4.1}$ & $59.7_{\pm13.4}$ & $45.4_{\pm5.6}$ & $75.0_{\pm5.3}$ & $55.9_{\pm2.6}$ & $60.4_{\pm2.6}$ \\
    \multicolumn{2}{c}{MMC-PCFG} & HT(2.4M) & $78.0_{\pm2.8}$ & $69.8_{\pm4.1}$ & $79.2_{\pm5.4}$ & $50.6_{\pm14.2}$ & $41.5_{\pm5.1}$ & $71.7_{\pm10.0}$ & $58.3_{\pm1.8}$ & $63.0_{\pm1.4}$ \\
    \multicolumn{2}{c}{\textbf{PTC-PCFG}} & HT(2.4M) & $\mathbf{81.9}_{\pm3.1}$ & $\mathit{71.5}_{\pm4.5}$ & $\mathit{83.0}_{\pm4.1}$ & $59.0_{\pm15.5}$ & $40.7_{\pm3.4}$ & $\mathbf{81.7}_{\pm6.2}$ & $\mathbf{61.1}_{\pm2.0}$ & $\mathbf{65.6}_{\pm1.7}$ \\
    \bottomrule
    
    \end{tabular}
    \label{tab:YouCook2}
\end{table*}

\begin{table*}[hbt!]
\small
\setlength{\tabcolsep}{4pt}
    \centering
    \caption{Performance comparison on MSRVTT.}
    \begin{tabular}{ccccccccccc}
    \toprule
	\multicolumn{2}{c}{Method} & Trainset & NP & VP & PP & SBAR & ADJP & ADVP & C-F1 & S-F1 \\
	\hline
    \multicolumn{2}{c}{LBranch} & None & $34.6$ & $0.1 $ & $0.9 $ & $0.2 $ & $3.8 $ & $0.3 $ & $14.4$ & $16.8$ \\
    \multicolumn{2}{c}{RBranch} & None & $34.6$ & $\mathbf{90.9}$ & $67.5$ & $\mathbf{94.8}$ & $25.4$ & $54.8$ & $54.2$ & $58.6$ \\
    \multicolumn{2}{c}{Random} & None & $34.6_{\pm0.1 }$ & $26.8_{\pm0.1}$ & $28.1_{\pm0.2 }$ & $24.6_{\pm0.3 }$ & $24.8_{\pm1.0}$ & $28.1_{\pm1.4}$ & $27.2_{\pm0.1}$ & $30.5_{\pm0.1}$ \\
    \multicolumn{2}{c}{C-PCFG} & MSRVTT & $46.6_{\pm3.2 }$ & $61.1_{\pm3.3}$ & $72.5_{\pm8.3 }$ & $63.7_{\pm4.0 }$ & $33.1_{\pm7.1}$ & $67.1_{\pm4.7}$ & $50.7_{\pm3.2}$ & $55.0_{\pm3.2}$ \\
    \multirow{11}{*}{\rotatebox{90}{VC-PCFG}} & ResNeXt & MSRVTT & $48.6_{\pm3.0 }$ & $59.0_{\pm6.0}$ & $72.0_{\pm3.6 }$ & $62.1_{\pm5.2 }$ & $32.6_{\pm2.5}$ & $70.4_{\pm6.4}$ & $50.7_{\pm1.7}$ & $54.9_{\pm2.2}$ \\
    & SENet & MSRVTT & $49.0_{\pm4.4 }$ & $63.5_{\pm6.4}$ & $71.7_{\pm4.8 }$ & $60.9_{\pm10.6}$ & $34.0_{\pm6.4}$ & $\mathit{74.1}_{\pm1.9}$ & $52.2_{\pm1.2}$ & $56.0_{\pm1.6}$ \\
    & I3D & MSRVTT & ${53.9}_{\pm10.5}$ & $63.2_{\pm9.1}$ & $73.7_{\pm2.9 }$ & $65.3_{\pm9.1 }$ & ${35.0}_{\pm6.8}$ & $73.8_{\pm4.1}$ & $54.5_{\pm1.6}$ & ${59.1}_{\pm1.7}$ \\
    & R2P1D & MSRVTT & $52.8_{\pm3.6 }$ & $63.3_{\pm4.6}$ & $73.1_{\pm10.1}$ & $66.9_{\pm2.0 }$ & $34.0_{\pm2.2}$ & $72.5_{\pm4.2}$ & $54.0_{\pm2.5}$ & $58.0_{\pm2.3}$ \\
    & S3DG & MSRVTT & $48.2_{\pm4.4 }$ & $60.4_{\pm3.9}$ & $71.4_{\pm6.4 }$ & $58.1_{\pm8.2 }$ & $25.3_{\pm2.2}$ & $61.8_{\pm8.4}$ & $50.7_{\pm3.2}$ & $54.7_{\pm2.9}$ \\
    & Scene & MSRVTT & $50.7_{\pm1.6 }$ & $65.0_{\pm4.7}$ & ${78.6}_{\pm3.6 }$ & $\mathit{67.3}_{\pm3.9 }$ & ${34.5}_{\pm4.6}$ & $71.7_{\pm1.8}$ & ${54.6}_{\pm1.5}$ & $58.4_{\pm1.3}$ \\
    & Audio & MSRVTT & $50.0_{\pm1.1 }$ & $63.7_{\pm6.1}$ & $72.7_{\pm3.0 }$ & $61.9_{\pm6.5 }$ & ${34.5}_{\pm2.3}$ & $68.0_{\pm5.9}$ & $52.8_{\pm1.3}$ & $56.7_{\pm1.4}$ \\
    & OCR & MSRVTT & $48.3_{\pm8.3 }$ & $57.1_{\pm4.6}$ & $\mathit{76.9}_{\pm0.6 }$ & $60.7_{\pm4.9 }$ & $33.9_{\pm8.3}$ & $72.1_{\pm4.4}$ & $51.0_{\pm3.0}$ & $55.5_{\pm3.0}$ \\
    & Face & MSRVTT & $46.5_{\pm6.8 }$ & $61.3_{\pm3.6}$ & $71.5_{\pm7.1 }$ & $60.8_{\pm11.0}$ & $30.9_{\pm3.4}$ & $68.4_{\pm6.0}$ & $50.5_{\pm2.6}$ & $54.5_{\pm2.6}$ \\
    & Speech & MSRVTT & $48.5_{\pm7.6 }$ & $60.7_{\pm3.5}$ & $74.5_{\pm5.7 }$ & $62.6_{\pm6.2 }$ & $27.3_{\pm1.8}$ & $74.0_{\pm3.1}$ & $51.7_{\pm2.6}$ & $56.2_{\pm2.5}$ \\
    & Concat & MSRVTT & $43.6_{\pm6.0 }$ & $64.7_{\pm3.0}$ & $68.5_{\pm8.0 }$ & $63.8_{\pm3.8 }$ & $32.0_{\pm5.5}$ & $70.4_{\pm5.9}$ & $49.8_{\pm4.1}$ & $54.2_{\pm4.0}$ \\
    \multicolumn{2}{c}{MMC-PCFG} & MSRVTT & ${52.3}_{\pm5.1 }$ & $\mathit{68.1}_{\pm2.9}$ & $\mathbf{78.2}_{\pm1.9 }$ & $65.8_{\pm2.4 }$ & $32.0_{\pm2.0}$ & $\mathbf{74.7}_{\pm2.3}$ & ${56.0}_{\pm1.4}$ & ${60.0}_{\pm1.2}$ \\
    \multicolumn{2}{c}{MMC-PCFG} & DiDeMo & $61.8_{\pm7.7}$ & $41.5_{\pm11.8}$ & $64.6_{\pm5.2}$ & $47.1_{\pm11.1}$ & $30.5_{\pm7.1}$ & $62.2_{\pm5.1}$ & $49.6_{\pm1.4}$ & $53.8_{\pm0.9}$ \\
    \multicolumn{2}{c}{MMC-PCFG} & YouCook2 & $40.7_{\pm14.9}$ & $23.9_{\pm3.4}$ & $59.9_{\pm10.2}$ & $16.2_{\pm2.6}$ & $14.5_{\pm4.0}$ & $23.7_{\pm3.9}$ & $34.0_{\pm6.4}$ & $37.5_{\pm6.8}$ \\
    \midrule

    \multicolumn{2}{c}{C-PCFG} & HT(29.6k) & $68.6_{\pm2.1}$ & $25.1_{\pm4.8}$ & $37.5_{\pm12.1}$ & $33.4_{\pm3.6}$ & $27.7_{\pm2.9}$ & $41.9_{\pm5.0}$ & $42.3_{\pm3.3}$ & $46.0_{\pm3.1}$ \\
    \multicolumn{2}{c}{MMC-PCFG} & HT(29.6k) & $70.8_{\pm2.7}$ & $32.2_{\pm13.2}$ & $48.6_{\pm6.3}$ & $36.0_{\pm4.6}$ & $32.0_{\pm2.1}$ & $43.1_{\pm5.5}$ & $47.2_{\pm3.8}$ & $51.7_{\pm5.0}$ \\
    \multicolumn{2}{c}{\textbf{PTC-PCFG}} & HT(29.6k) & $62.2_{\pm7.7}$ & $54.0_{\pm13.0}$ & $60.0_{\pm4.9}$ & $53.0_{\pm3.6}$ & $32.4_{\pm4.0}$ & $45.9_{\pm5.1}$ & $52.2_{\pm3.9}$ & $57.4_{\pm5.1}$ \\
    \midrule

    \multicolumn{2}{c}{C-PCFG} & HT(296k) & $75.5_{\pm1.4}$ & $34.8_{\pm4.3}$ & $58.6_{\pm1.6}$ & $46.9_{\pm2.8}$ & $40.0_{\pm3.1}$ & $55.4_{\pm7.1}$ & $52.0_{\pm1.3}$ & $56.4_{\pm1.7}$ \\
    \multicolumn{2}{c}{MMC-PCFG} & HT(296k) & $75.1_{\pm2.5}$ & $39.4_{\pm15.5}$ & $55.2_{\pm7.2}$ & $40.0_{\pm4.5}$ & $40.2_{\pm5.8}$ & $51.0_{\pm6.4}$ & $52.4_{\pm5.5}$ & $56.8_{\pm6.4}$ \\
    \multicolumn{2}{c}{\textbf{PTC-PCFG}} & HT(296k) & $70.2_{\pm5.8}$ & $51.7_{\pm12.1}$ & $64.5_{\pm6.2}$ & $54.0_{\pm6.5}$ & $39.2_{\pm3.2}$ & $54.8_{\pm9.5}$ & $55.7_{\pm5.0}$ & $61.1_{\pm5.9}$ \\
    \midrule

    \multicolumn{2}{c}{C-PCFG} & HT(592k) & $76.9_{\pm2.6}$ & $35.4_{\pm8.3}$ & $57.9_{\pm5.4}$ & $44.5_{\pm10.2}$ & $\mathbf{41.4}_{\pm3.9}$ & $57.8_{\pm5.3}$ & $52.5_{\pm3.4}$ & $56.4_{\pm3.6}$ \\
    \multicolumn{2}{c}{MMC-PCFG} & HT(592k) & $76.1_{\pm3.4}$ & $46.3_{\pm20.0}$ & $57.0_{\pm5.8}$ & $50.1_{\pm10.1}$ & $37.9_{\pm3.1}$ & $52.8_{\pm4.6}$ & $55.1_{\pm7.0}$ & $60.2_{\pm8.0}$ \\
    \multicolumn{2}{c}{\textbf{PTC-PCFG}} & HT(592k) & $74.0_{\pm3.6}$ & $50.2_{\pm18.9}$ & $67.0_{\pm4.1}$ & $54.5_{\pm9.1}$ & $34.7_{\pm2.4}$ & $55.4_{\pm7.3}$ & $\mathbf{57.4}_{\pm4.6}$ & $\mathbf{62.8}_{\pm5.7}$ \\
    \midrule

    \multicolumn{2}{c}{C-PCFG} & HT(1.2M) & $77.0_{\pm2.0}$ & $30.5_{\pm10.4}$ & $60.1_{\pm3.4}$ & $41.5_{\pm11.1}$ & $38.5_{\pm4.4}$ & $52.2_{\pm4.8}$ & $51.6_{\pm3.1}$ & $55.5_{\pm3.5}$ \\
    \multicolumn{2}{c}{MMC-PCFG} & HT(1.2M) & $\mathbf{77.9}_{\pm2.2}$ & $44.2_{\pm13.1}$ & $40.6_{\pm22.2}$ & $45.6_{\pm5.5}$ & $\mathit{40.5}_{\pm4.6}$ & $56.3_{\pm6.8}$ & $52.4_{\pm4.5}$ & $57.2_{\pm5.1}$ \\
    \multicolumn{2}{c}{\textbf{PTC-PCFG}} & HT(1.2M) & $72.3_{\pm1.6}$ & $44.0_{\pm7.7}$ & $70.2_{\pm4.6}$ & $53.4_{\pm7.9}$ & $34.4_{\pm4.6}$ & $57.4_{\pm6.9}$ & $55.6_{\pm2.5}$ & $61.0_{\pm3.3}$ \\
    \midrule

    \multicolumn{2}{c}{C-PCFG} & HT(2.4M) & $\mathit{77.2}_{\pm2.1}$ & $31.1_{\pm4.6}$ & $59.8_{\pm8.0}$ & $43.3_{\pm9.7}$ & $39.1_{\pm3.5}$ & $54.5_{\pm6.4}$ & $51.9_{\pm2.3}$ & $55.4_{\pm2.8}$ \\
    \multicolumn{2}{c}{MMC-PCFG} & HT(2.4M) & $76.2_{\pm2.2}$ & $36.1_{\pm6.3}$ & $62.5_{\pm7.5}$ & $47.8_{\pm8.1}$ & $40.0_{\pm4.5}$ & $55.8_{\pm5.3}$ & $53.5_{\pm2.4}$ & $57.1_{\pm2.8}$ \\
    \multicolumn{2}{c}{\textbf{PTC-PCFG}} & HT(2.4M) & $74.2_{\pm2.8}$ & $46.0_{\pm7.1}$ & $67.5_{\pm4.1}$ & $52.7_{\pm8.6}$ & $40.2_{\pm5.6}$ & $58.3_{\pm10.7}$ & $\mathit{56.6}_{\pm2.5}$ & $\mathit{61.2}_{\pm2.9}$ \\
    \bottomrule
    \end{tabular}
    \label{tab:MSRVTT}
\end{table*}

\clearpage

\end{document}